\documentclass{article}

    \PassOptionsToPackage{numbers, compress}{natbib}

    \usepackage[preprint]{neurips_2024}

\usepackage[utf8]{inputenc} 
\usepackage[T1]{fontenc}    
\usepackage{hyperref}       
\usepackage{url}            
\usepackage{booktabs}       
\usepackage{amsfonts}       
\usepackage{nicefrac}       
\usepackage{microtype}      
\usepackage{xcolor}         
\usepackage{graphicx}
\usepackage{multirow}
\usepackage{booktabs}
\usepackage{amssymb} 
\usepackage{wrapfig}
\usepackage{subcaption}
\usepackage{amsmath}

\title{Your Network May Need to Be Rewritten: Network Adversarial Based on High-Dimensional Function Graph Decomposition}

\author{%
  Xiaoyan Su\thanks{Joint first authors.} \ \thanks{Corresponding authors.} \\
  Shenzhen University\\
  \texttt{2021022023@email.szu.edu.cn} \\
  \And
  Yinghao Zhu\footnotemark[1] \ \footnotemark[2] \\
  Alibaba Digital Media \& Entertainment Group \\
  \texttt{22160083@zju.edu.cn} \\
  \And
  Run Li \\
  Lanzhou University of Technology \\
  \texttt{lirun4977@gmail.com} \\
}

\begin{document}

\maketitle

\begin{abstract}
In the past, research on a single low dimensional activation function in networks has led to internal covariate shift and gradient deviation problems. A relatively small research area is how to use function combinations to provide property completion for a single activation function application. We propose a network adversarial method to address the aforementioned challenges. This is the first method to use different activation functions in a network. Based on the existing activation functions in the current network, an adversarial function with opposite derivative image properties is constructed, and the two are alternately used as activation functions for different network layers. For complex situations, we propose a method of high-dimensional function graph decomposition(HD-FGD), which divides it into different parts and then passes through a linear layer. After integrating the inverse of the partial derivatives of each decomposed term, we obtain its adversarial function by referring to the computational rules of the decomposition process. The use of network adversarial methods or the use of HD-FGD alone can effectively replace the traditional MLP+activation function mode. Through the above methods, we have achieved a substantial improvement over standard activation functions regarding both training efficiency and predictive accuracy. The article addresses the adversarial issues associated with several prevalent activation functions, presenting alternatives that can be seamlessly integrated into existing models without any adverse effects. We will release the code as open source after the conference review process is completed.
\end{abstract}

\section{Introduction}
With the explosion of LLM, deep learning has once again become an important research topic in many scientific fields. Both LLM and various neural network models rely on the nonlinearity introduced by activation functions, and using appropriate activation functions can represent and learn more complex functional relationships. In order to enhance the nonlinear expression ability of the model, many non parametric activation functions that are suitable for specific tasks have been proposed according to the different application tasks. The currently popular activation functions include Sigmoid, Tanh, ReLu, GeLu \cite{hendrycks2016gaussian}, Swish \cite{ramachandran2017searching}, etc. Regardless of the activation function, it is limited to introducing nonlinearity in a single two-dimensional space, inevitably causing changes in the distribution of layer inputs, leading to internal covariate shift (ICS) \cite{ioffe2015batch} and gradient deviation problems.

In order to alleviate ICS as much as possible during model training, only the same activation function is considered at different network layers to increase model nonlinearity. As the network depth increases, the gradient may have varying degrees of deviation, which leads to ineffective training \cite{dubey2022activation}. Therefore, we consider whether it is possible to use an activation function to generate ICS in the model, while also having a non-linear adversarial function to correct the gradient, pulling the trained data distribution back to the original data distribution, so that the gradient can be smoothly transmitted back.

\textbf{Our Contributions.} Given the challenges posed by introducing activation functions in the network, this paper proposes a network adversarial method. For complex activation functions, the adversarial function is solved through high-dimensional function graph decomposition (HD-FGD). The partial derivative of the function is derived from the derivative of the original activation function. Although the image properties are opposite, it does not introduce additional nonlinearity to the model and cause overfitting. By cross using the original activation function and its adversarial function in the network layer of the model to activate the network, gradient updates are made smooth during backpropagation. Our main contributions are in three aspects as follows

\begin{itemize}
    \item Propose a universal network adversarial method and incorporate it into the model system, which can effectively solve the problems of ICS and gradient deviation during the training process, while significantly improving the predictive performance of the model.
    \item For the solution of adversarial functions for complex activation functions, we propose a method based on HD-FGD to simplify the process. Users split it into different expressions according to their wishes. After verification, the split activation function performs significantly better than the original activation function, providing a new idea for the transformation of two-dimensional activation functions into high-dimensional research.
    \item The effectiveness of network adversarial method has been demonstrated, and the adversarial functions of common activation functions have been solved. The combination of the corresponding activation functions and their adversarial functions can replace the corresponding activation functions in all linear layers of the model without loss, and significantly accelerate the convergence speed of the model.
\end{itemize}

For tasks in different fields, our method was validated in the network layers of multiple models. Experiments have shown that compared to the original activation function, using network adversarial in the network layers of each model can significantly improve model performance and training speed. The activation function obtained by using HD-FGD can effectively replace the original activation function, especially in the linearity layer, with significant effects.

\section{Related Work}
\subsection{The Evolution of Nonlinear Learning}
Early neural networks introduced nonlinearity through non parametric activation functions such as Sigmoid and ReLu \cite{eger2019time, liu2019natural, basirat2020relu, gu2019fast}. With the widespread application of convolutional networks, new activation functions such as ELU \cite{clevert2020fast, trottier2017parametric} aim to achieve higher performance cross structures in models. Although the effect is significant, it relies more on architecture and task attributes. In order to achieve adjustable nonlinear expression of the model during training, an adaptive activation function APL \cite{agostinelli2014learning} with learnable parameters is proposed, which performs excellently in multi-scale tasks. At the same time, there are also studies attempting to make the activation function learnable to adapt to different task attributes. For example, Zoph et al. \cite{zoph2016neural} used reinforcement learning to train multiple networks and generated nonlinearity in the results. Although this method generated more generalized activation functions, it cannot guarantee the existence of the optimal generalizable function in strict space. Hyperactivations \cite{vercellino2017hyperactivations} innovatively explores the search space of activation functions and uses hypernetworks to generate weights for the activation network, effectively improving network training speed and predictive performance.

Overall, in the past, most activation functions with different properties were suitable for certain tasks in specific fields. Nowadays, the nonlinear design of activation functions tends to be biased towards mining potential complex relationships in specific task data, especially Hyperactivations \cite{vercellino2017hyperactivations}, which uses a nested activation network with generative weight networks, providing a new paradigm for improving the expression ability of the required attributes of activation functions.

\subsection{ICS and Gradient Deviation}
The promotion of deep learning benefits from the progress of many algorithms and architectures, such as Dropout \cite{srivastava2014dropout}, initial weight initialization methods \cite{sutskever2013importance, he2016identity, klambauer2017self, he2015delving}, optimizers \cite{diederik2014adam, zhuang2020adabelief, luo2019adaptive}, etc. Among them, the most prominent example is batch normalization \cite{ioffe2015batch}, which improves neural networks by introducing additional network layers to control the mean and variance of layer distributions, and can solve the problem of ICS from the dimension of data distribution correction. Based on this, the layer normalization method \cite{ba2016layer} has been proposed, which is not affected by batch size and is more suitable for processing sequential data tasks such as natural language processing. In addition, there is also group normalization technique \cite{wu2018group}, which divides each sample feature into multiple groups and normalizes them within each group, reducing dependence on global information and memory overhead. Other normalization based algorithms \cite{li2018adaptive, shi2016real} have also been proposed. SELU \cite{klambauer2017self} uses a weight initialization method based on linear transformation to improve the training stability of the network, but its hyperparameters are affected by human intervention and have higher requirements for the distribution of input data, making it more suitable for specific fields. In Mattia J. Villani's work \cite{jiang2022delve}, it was demonstrated that deep neural networks obtained through the ReLU function exhibit a certain degree of degradation. Specific methods can be used to express deep networks using shallow networks, further simplifying the network structure and reducing the possibility of gradient shift.

In summary, the above contributions have paved the way for the research of deep learning in various fields and provided new ideas for the subsequent ICS and gradient shift problems, such as enabling the activation function to have adaptive mechanisms during training.

\section{Network Adversarial Method}
In order to better stabilize the gradient improvement model prediction ability during the training process and minimize ICS, we propose a network adversarial method, which is divided into global adversarial (GA) and split adversarial (SA). The HD-FGD in SA effectively replaces the original activation function, while proposing the idea of mapping two-dimensional activation functions to high-dimensional space, providing new exploration space for the research of deep learning in this part.

\subsection{Verification of Gradient Adversarial Thinking}
Assuming a k-layer deep neural network is used for classification tasks, and all linear layer activation functions are Sigmoid, $F(x)$ represents its derivative function, $x$ is the feature data of the input model, $y$ is the true label, $\hat y$ is the model prediction result, $w_i$ is the parameter of the i-th layer, $x_i$ is the input data of the i-th layer, $l$ is the loss function, and the gradient $\lambda_i$ of each layer can be expressed as
\begin{equation}\label{f1}
\setlength{\arraycolsep}{2pt} 
\renewcommand{\arraystretch}{2} 
\begin{array}{l}
 \lambda _i  = \frac{{\partial l}}{{\partial w_i }} = \frac{{\partial l}}{{\partial \hat y}} \cdot \frac{{\partial \hat y}}{{\partial w_i }} \\ 
  = \frac{{\partial l}}{{\partial \hat y}} \cdot \frac{{\partial (x_k w_k )}}{{\partial x_k }} \cdot \frac{{\partial (x_{k - 1} w_{k - 1} )}}{{\partial x_{k - 1} }} \cdot \frac{{\partial (x_{k - 2} w_{k - 2} )}}{{\partial x_{k - 2} }} \ldots \frac{{\partial (x_1 w_1 )}}{{\partial w_1 }} \\ 
  = \frac{{\partial l}}{{\partial \hat y}} \cdot w_k F(x_k ) \cdot w_{k - 1} F(x_{k - 1} ) \cdot w_{k - 2} F(x_{k - 2} ) \ldots x_i  \\ 
  = \frac{{\partial l}}{{\partial \hat y}} \cdot (\prod\limits_{j = 1}^{k - i} {w_{k - j + 1}  \cdot F(x_{k - j + 1} )} ) \cdot x_i  \\ 
\end{array}
\end{equation}

The gradient of each layer is influenced by two parts: the multiplication between the parameters of each previous layer and the gradient of the input data for each layer, and the other part is the input data for that layer. In the work of Sergey et al. \cite{ioffe2015batch}, the use of batch normalization effectively controls the stability of the input distribution in the control layer, and we assume that it approximates a normal distribution, The derivative interval of Sigmoid is between 0 and 0.25. As the network deepens, excessive multiplication will cause the gradient to disappear, while multiplication with a derivative interval greater than 1 will cause the gradient to explode. For formula (\ref{f1}), some of the formulas can be split into pairwise groups.
\begin{equation}\label{f2}
\lambda _i  = \frac{{\partial l}}{{\partial \hat y}} \cdot (\prod\limits_{j = 1}^{(k - i)/2} {w_{k - (2j - 1) + 1}  \cdot \underbrace {F(x_{k - (2j - 1) + 1} )}_A}  \cdot w_{k - (2j - 1)}  \cdot \underbrace {F(x_{k - (2j - 1)} )}_B) \cdot x_i 
\end{equation}

The gradient stability interval of a single activation function is limited. If $A$ and $B$ in formula (\ref{f2}) can generate gradient pulling and have gradient stability intervals respectively, there is a greater possibility that gradient deviation will not occur during model training. The two are alternately used as activation functions for different network layers during backpropagation, forming gradient adversaries to maintain a stable state of the gradient. Compared with the study of a single activation function, we study the combination of two or more activation functions to flexibly control the required properties of the activation function, reduce the frequency of gradient deviation and the amplitude of ICS. This is the main idea of network adversarial method.

To validate our idea, the following experiment is proposed. We construct a multi-layer linear network and add batchnorm layers while using Sigmoid activation as the control group. The experimental group, on the basis of the above, uses Sigmoid and its adversarial function $\xi _{{\rm{Sigmoid}}} (x)$ alternately as the activation function. The following image shows the data distribution, gradient distribution, and predictive performance after passing through different linear layers, and these results further validate the feasibility of our idea.

\begin{figure}
  \centering
  \begin{minipage}[b]{0.32\textwidth}
    \centering
    \includegraphics[width=\textwidth]{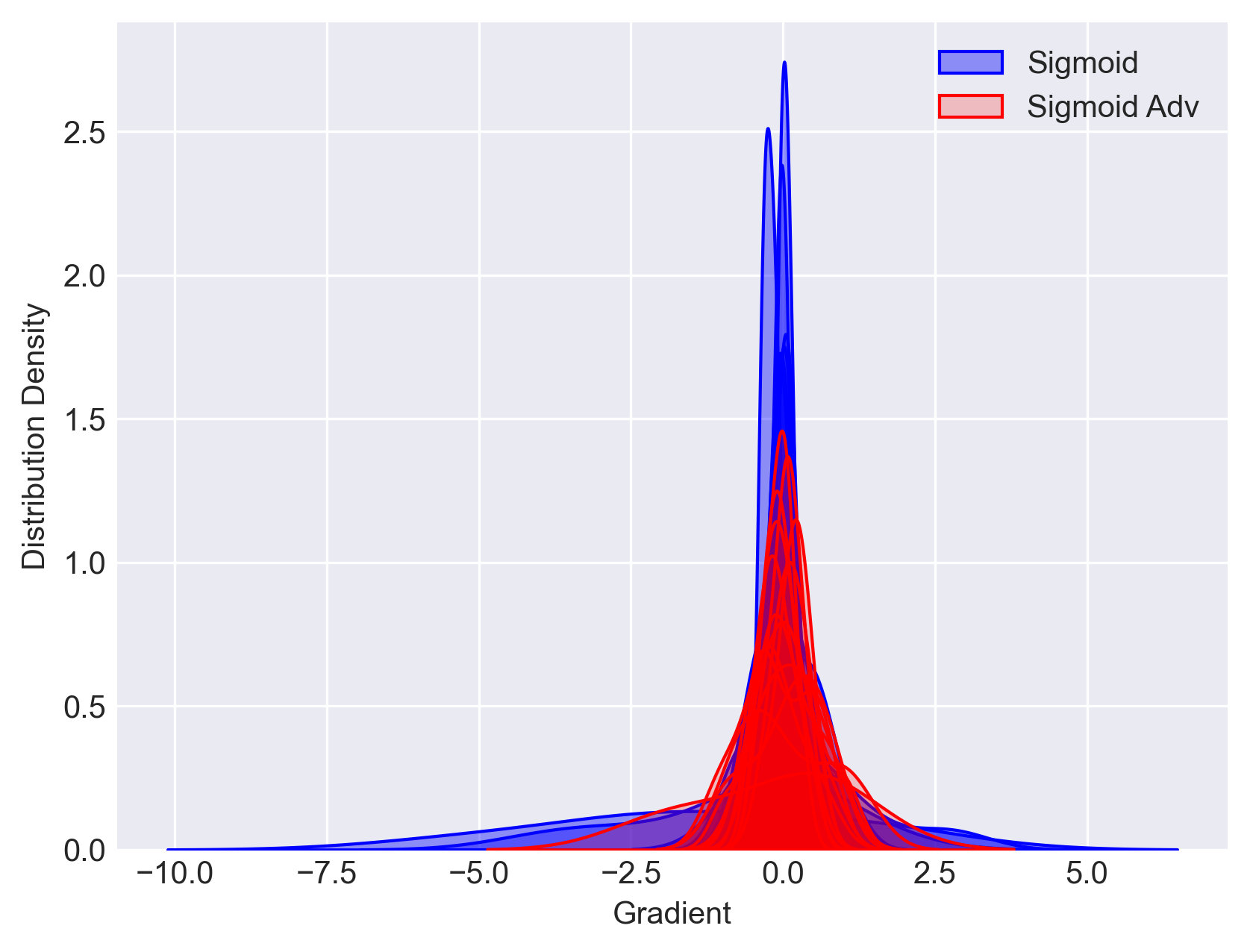}
    \subcaption{(a) Gradient distribution}
  \end{minipage}
  \hfill
  \begin{minipage}[b]{0.32\textwidth}
    \centering
    \includegraphics[width=\textwidth]{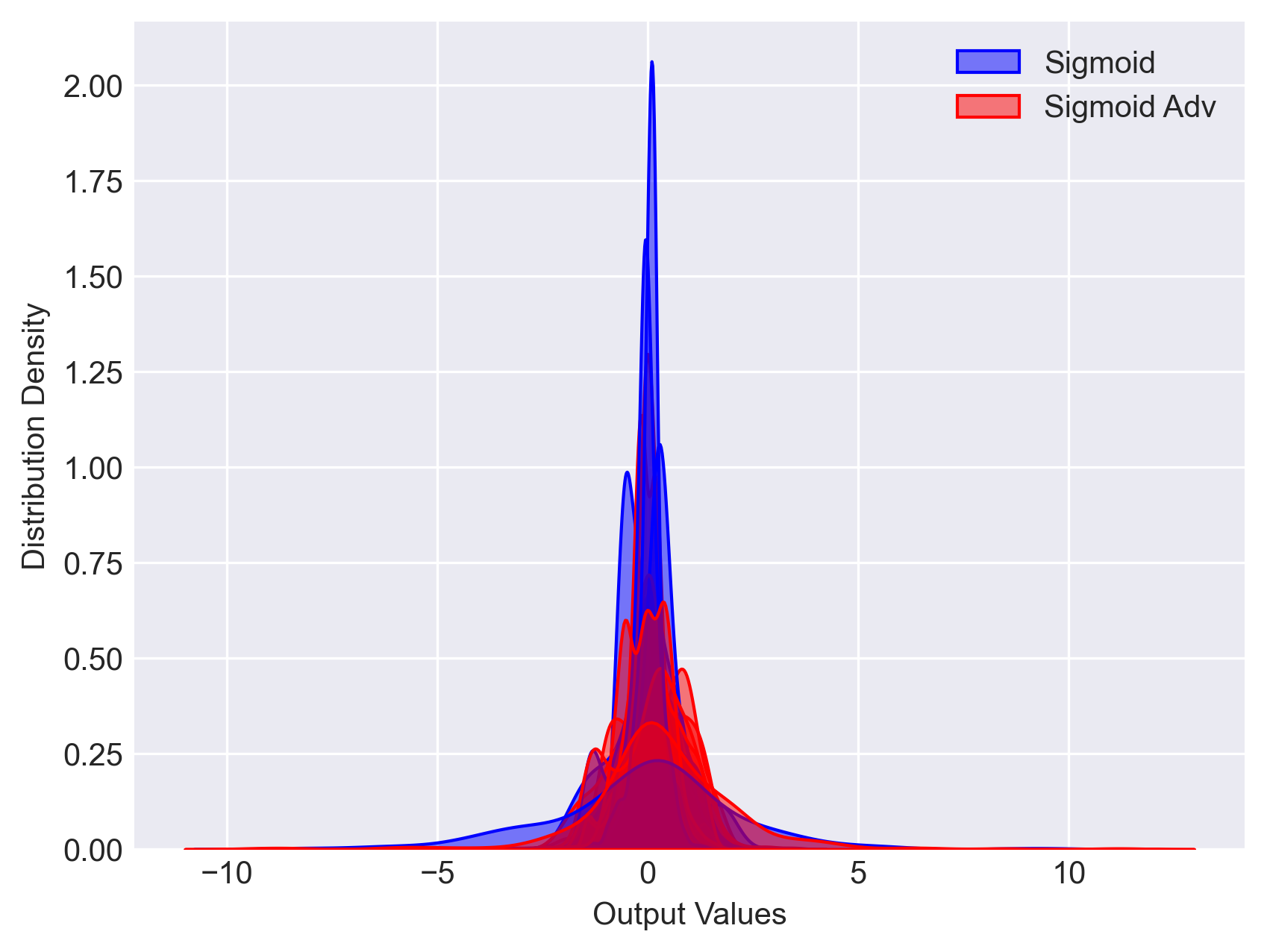}
    \subcaption{(b) Data distribution}
  \end{minipage}
  \hfill
  \begin{minipage}[b]{0.32\textwidth}
    \centering
    \includegraphics[width=\textwidth]{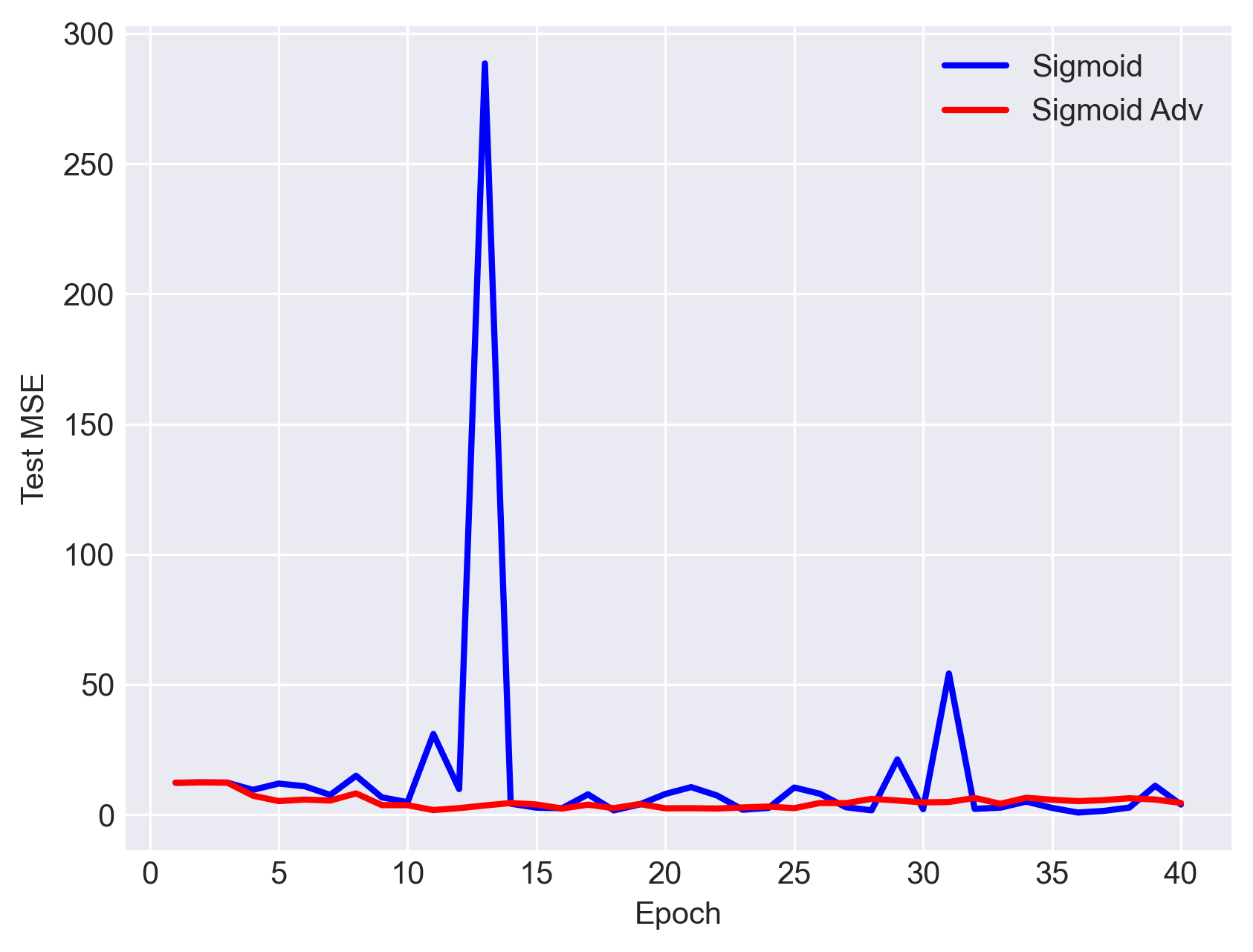}
    \subcaption{(c) Predictive performance}
  \end{minipage}
  \caption{Measure ICS and gradient deviation in networks with and without network adversarial methods. Sigmoid-Adv in the figure is $\xi _{{\rm{Sigmoid}}} (x)$. According to the changes in gradient distribution (a), it can be seen that the linear layer gradient distribution with the addition of network adversarial method is more uniform, and the distribution of extreme gradients is significantly improved. The data distribution of each linear layer in the experimental group (b) fluctuates significantly less than that in the control group. In terms of performance (c), the MSE of the experimental group maintains a significant advantage in all iterations, and there is no significant fluctuation.}
  \label{fig:1}
\end{figure}

\subsection{The Proposal of HD-FGD}
We will provide a general explanation of the network adversarial method proposed in this article. Specifically, for simple activation functions such as Sigmoid, we can directly perform GA operations. For complex activation functions such as Tanh and GeLu, we need to use SA operations. Otherwise, there may be difficulties in solving or exponential growth of adversarial gradients. Taking Tanh as an example, the adversarial function \(\frac{{e^{4x}  - 1}}{{8e^{2x} }} + \frac{x}{2} + C\) is solved using GA. During the backpropagation process, the adversarial gradient is more prone to gradient deviation due to the risk of exponential explosion, leading to training termination. The SA based on HD-FGD is to represent complex functions with multiple simple functions, and provide a trainable network for different splitting terms to achieve the transformation from low dimensional functions to high dimensions, simplifying the solving process while ensuring stable training.
\begin{figure}
\centering
  \includegraphics[width=0.94\textwidth]{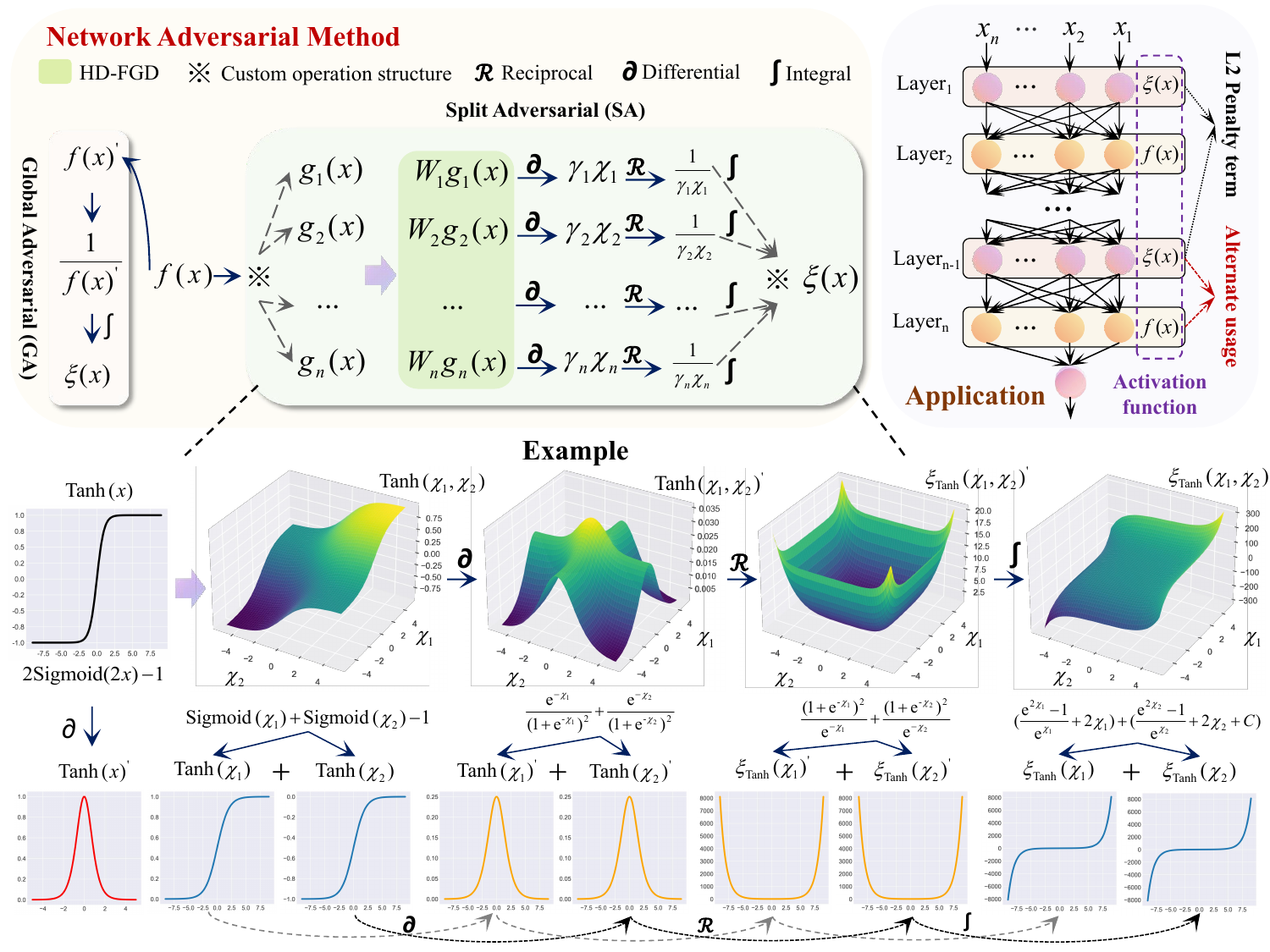}
  \caption{Schematic diagram of network adversarial method.}
  \label{fig:2}
\end{figure}

 As shown in Figure \ref{fig:2}, network adversarial method can be divided into two forms: GA and SA. For activation function $f(x)$, when using GA, the reciprocal of its derivative is the derivative $\xi(x)^{'}$ of adversarial function $\xi(x)$. After integrating $\xi(x)^{'}$, $\xi(x)$ is obtained.

For the scaling mapping $\chi  = AF\left({Wx}\right)$ of a single linear layer of data input, this process can essentially be seen as adding a trainable matrix $W$ to any activation function $f(x)$, which can be represented as $f(W,x)$, where $AF\left(  \cdot  \right)$ represents the activation function. Therefore, it can be considered to replace this linear mapping with multiple linear mappings $f(\chi _1 ,\chi _2  \ldots \chi _n )$, each of which introduces nonlinearity with different properties to the model, making the expression of the activation function more flexible. This splitting can reduce the number of hidden units h in the original trainable parameter matrix to $n$ trainable parameter matrices with $h/n$ hidden units, and optimize the computing speed of $W_i$ using parallel operations.As shown in Figure \ref{fig:3}, under normal circumstances, the non-linear introduction of the model is represented by $x \cdot W \to f(x) \to x'$, where $x'$ represents the output after linear mapping and activation function, while HD-FGD optimizes the process of $W \to f(x)$, while transforming two-dimensional functions into high-dimensional space.

When using SA, first convert the original activation function into a linearly mapped expression $f(x) = f(W,x)$, the original activation function is decomposed into a computational structure composed of basic operations using  HD-FGD. In the simplest case, if all basic operation symbols are the same, the original activation function can be represented as the accumulation $\sum\limits_{i = 1}^n {W_i g_i (x)}$ or multiplication term of each splitting term $\prod\limits_{i = 1}^n {W_i g_i (x)}$, where $g_i(x)$ is the different parts of the original activation function after splitting. For the split $g_i(x)$, set their respective trainable parameter matrices $W_i$, where $n$ represents the specific number of split terms, which is determined by the specific expression of the activation function or can be artificially split into a specified value. When splitting, it is best to use methods such as \cite{ioffe2015batch} to process the input data as a zero mean, while ensuring that there is a saturation interval in the value range of $g_i(x)$. Otherwise, the solved adversarial function needs to be truncated using clip \cite{bungert2021clip} or clamp \cite{pascanu2013difficulty}. Following the above requirements can accelerate network convergence and reduce the risk of gradient deviation. According to $\chi  = AF\left({Wx}\right)$, there is $f(x) = \sum\limits_{i = 1}^n {\varphi (\chi _i )}$, where $\varphi (\chi _i )$ represents the i-th splitting term after passing through the linear layer, and the sum of partial derivatives of $f(x)$ can be expressed as $\sum\limits_{i = 1}^{\rm{n}} {\gamma _i  \cdot \chi _i }$, where $\gamma _i$ is a constant term. The sum of partial derivatives of $\xi (x)$ can be expressed as $\xi (\chi _1 ,\chi _1  \ldots \chi _{\rm{n}} )^{'}  = \sum\limits_{i = 1}^{\rm{n}} {\frac{1}{{\gamma _i \chi _i }}} $, and by integrating it, the adversarial function $\xi (x) = \xi (\chi _1 ,\chi _2  \ldots \chi _{\rm{n}} ) = \sum\limits_{i = 1}^{\rm{n}} {\int {\frac{1}{{\gamma _i \chi _i }}} } {\rm{d}}\chi _i$ can be obtained. When using the adversarial function as the activation function, in order to avoid overfitting caused by too many splitting terms, we add an L2 penalty term in gradient backpropagation as a constraint.

Taking Tanh as an example in Figure \ref{fig:2}, it can be transformed into an expression for Sigmoid
\begin{equation}\label{f3}
{\rm{Tanh}}(x) = \frac{{e^x  - e^{ - x} }}{{e^x  + e^{ - x} }} = \frac{{2e^x  - e^x  - e^{ - x} }}{{e^x  + e^{ - x} }} = \frac{2}{{1 + e^{ - 2x} }} - 1 = 2{\rm{Sigmoid}}(2x) - 1
\end{equation}
Using a SA based on HD-FGD to solve its adversarial function, it can be divided into two parts, each of which passes through a linear mapping network $W_i$. Let $\chi _i  = W_i x$, then Tanh can be expressed as
\begin{equation}\label{f4}
{\rm{Tanh(}}x{\rm{)}} = {\rm{Tanh}}(\chi _1 ,\chi _2 ) = {\rm{Sigmoid}}(\chi _1 ) + {\rm{Sigmoid}}(\chi _2 ) - 1
\end{equation}

At this point, the operation realizes the mapping of two-dimensional Tanh to high-dimensional space, so the study of Tanh's derivative can also shift from two-dimensional to three-dimensional space. The derivative of its adversarial function is the sum of the reciprocal of the partial derivatives of ${\rm{Tanh}}(\chi _1 ,\chi _2 )$, and after integration, it becomes Tanh's adversarial function.

Based on experimental verification, we have gained some experiences that can make network adversarial method more effective:

\textbf{Small gradients are effective in the end.} For any neural network model, we alternately use the original activation function and its adversarial function as the activation function for the corresponding layer in the linear layer. In backpropagation, deep network gradients affect shallow network gradients, so the activation function gradient in the latter layer can counteract the gradient in the previous layer, but the latter layer can only rely on deeper gradients to counteract it. Taking Sigmoid as an example, if the activation function of the last layer of the network is its adversarial function, its gradient will not have any small gradients against it, which is prone to gradient explosion. Therefore, the activation function of the latter layer of the network should be as conservative as possible, and choosing a small gradient activation function in the latter layer will improve the performance of the network.

\textbf{Better results are achieved when used in conjunction with batch normalization.} When batch normalization is applied to the input network data, most gradients are well constrained within the stable range of the activation function. For Sigmoid, it avoids gradient disappearance caused by extreme inputs, and the derivative symmetric activation function can better handle extreme inputs. 

\textbf{The activation function's derivative being symmetric about the y-axis leads to better performance.} If the gradient at $x_0$ of a certain network layer is $f(x_0 )^{'}$, the presence of $\xi (x)^{'} $ in $x \in [x_0  - \tau ,x_0  + \tau ]$ in the next network layer of backpropagation can pull $f(x_0 )'$ to a stable interval, and $\tau $ represents the allowable deviation range. At this time, $x \in [ x_0  - \tau, x_0  + \tau]$ is called the stable interval of the function combination of $x$ at $x_0$. At this point, if $f(x)'$ is symmetric about the y-axis, due to the opposite nature of the image between $f(x)'$ and $\xi (x)^{'} $, $\xi (x)^{'} $ is also symmetric about the y-axis. Furthermore, there exists a situation where $x \in [ - (x_0  - \tau ), - (x_0  + \tau )]$ also pulls $f(x)'$ back to the stable interval. Therefore, compared to asymmetric functions, the stable interval of the function combination is doubled.

\section{Application Instance}
This section takes Sigmoid as an example and uses GA to solve its adversarial function as a whole. Taking Tanh and GeLu (refer to the Appendix \ref{A-2-2}  for the derivation process) as examples, it shows the process of using HD-FGD to separate them into four terms and using SA to solve their adversarial functions separately.

\subsection{The Adversarial Function of Sigmoid} \label{4-1}
Using the GA of the network adversarial method to solve the adversarial function of Sigmoid, the derivation process is as follows
\begin{equation}\label{f5}
{\rm{Sigmoid}}(x) = \frac{{\rm{1}}}{{{\rm{1}} + e^{ - x} }} \Rightarrow {\rm{Sigmoid}}(x)^{'}  = \frac{{e^x }}{{{\rm{(1}} + e^x )^2 }} \Rightarrow \xi _{{\rm{Sigmoid}}} (x)^{'}  = \frac{{{\rm{(1}} + e^x )^2 }}{{e^x }}
\end{equation}
Among them, $\xi _{{\rm{Sigmoid}}} (x)^{'}$ represents the derivative of the Sigmoid adversarial function, which is integrated to obtain the adversarial function $\xi _{{\rm{Sigmoid}}} (x)$. Using the substitution method, define $u = e^{x} $, then ${\rm{d}}u = e^{x} {\rm{d}}x$, and the integration of $\xi _{{\rm{Sigmoid}}} (x)^{'}$ (derivation can be found in Appendix \ref{A-1-1}) is as follows
\begin{equation}\label{f6}
\int {\frac{{{\rm{(1}} + e^x )^2 }}{{e^x }}} {\rm{d}}x = \frac{{e^{2x}  - 1}}{{e^x }} + 2x + C
\end{equation}

In deep network training, even if ReLu grows at a rate of $y=x$, gradient deviation may occur. For Sigmoid and its adversarial functions, gradient deviation is mainly caused by the uncontrollability of network layer parameters under extreme input conditions. To address this issue, the derivative of the Sigmoid can be raised by $\alpha$ units on the original basis, with its derivative becoming $\frac{{e^x }}{{(1 + e^x )^2 }} + \alpha $ and the value interval of the derivative becoming $\left( {\alpha ,\alpha  + 0.25} \right)$. At this point, the derivative of the adversarial function is reciprocal to the function after the Sigmoid derivative is raised, and its derivative becomes $\frac{{(1 + e^x )^2 }}{{e^x  + \alpha (1 + e^x )^2 }}$. Therefore, the value interval of the derivative of the adversarial function becomes $\left(\frac{1}{{0.25 + \alpha }})\right)$, $\frac{1}{\alpha }$. During training, the parameter $\alpha$ is set as an adjustable parameter, and in this experiment, it is set to 1. When using the Sigmoid function and its adversarial function alternately, the gradient shift generated in a certain layer can be appropriately scaled by the value range of the adversarial function in the next layer, thereby correcting the shift. Due to the limitation of $\alpha$, gradient deviation is avoided in the gradient correction of the derivative of the adversarial function. The modified Sigmoid function ${\rm{sigmoid}}_\vartheta $ can be found in formula (\ref{f7}), and its adversarial function $\xi _{{\rm{sigmoid}}_\vartheta  } (x)$ (solution process can be found in Appendix \ref{A-1-2}) can be found in formula (\ref{f8}).
\begin{equation}\label{f7}
{\rm{Sigmoid}}_\vartheta  (x) = \frac{1}{{1 + e^{ - x} }} + \alpha x + C
\end{equation}

\begin{equation}\label{f8}
\xi _{{\rm{Sigmoid}}_\vartheta  } (x) = \frac{{{\rm{ln}}(2\alpha e^x  + \sqrt {4\alpha  + 1}  + 2\alpha  + 1) - {\rm{ln}}(\left| {2\alpha e^x  - \sqrt {4\alpha  + 1}  + 2\alpha  + 1} \right|)}}{{\alpha \sqrt {4\alpha  + 1} }} + \frac{x}{\alpha } + C
\end{equation}
Among them, $\alpha$ in formula (\ref{f8}) is greater than 0.

\subsection{The Adversarial Function of Tanh}
The expression of Tanh is $\frac{{e^x  - e^{ - x} }}{{e^x  + e^{ - x} }}$, and its adversarial function is solved using the SA in the network 
\begin{wrapfigure}{r}{0.5\textwidth}
  \centering
  \includegraphics[width=0.5\textwidth]{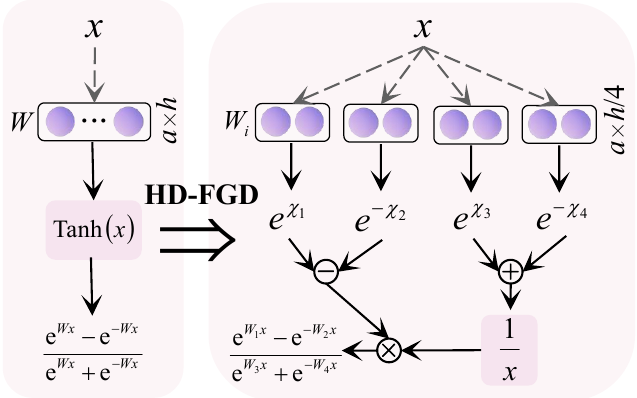}
  \caption{HD-FGD splits Tanh into four terms.}
  \label{fig:3}
\end{wrapfigure}
adversarial method. Due to differences in expression or the number of split terms, the split form of the same function does not have uniqueness. In addition to the 2-term splitting method in Figure \ref{fig:2}, Tanh can also be split into 4 terms. Firstly, add a linear layer to the Tanh function, and the network can be expressed as $\frac{{e^{Wx}  - e^{ - Wx} }}{{e^{Wx}  + e^{ - Wx} }}$, where $W$ represents the trainable parameter matrix. After specifying the number of split terms as $n=4$, a separate trainable parameter matrix $W_i$ can be added for each split part. Instead of learning a complex function relationship, $W$ now learns $n$ simple function relationships. Since each $W_i$ is used to learn the corresponding function relationship, the number of hidden units in $W_i$ can be moderately reduced. We define the number of hidden units for each split part in the linear layer as $\frac{1}{n}$. This process can be illustrated as shown in Figure \ref{fig:3}.

The split Tanh can be expressed as
\begin{equation}\label{f9}
{\rm{Tanh}}\left( {W_1 ,W_2 ,W_3 ,W_4 ,x} \right) = \frac{{e^{W_1 x}  - e^{W_2 x} }}{{e^{W_3 x}  + e^{W_4 x} }}
\end{equation}
Let $\chi _i  = W_i x$, then formula (\ref{f9}) can be modified to
\[
{\rm{Tanh}}\left( {\chi _1 ,\chi _2 ,\chi _3 ,\chi _4 } \right) = \frac{{e^{\chi _1 }  - e^{\chi _2 } }}{{e^{\chi _3 }  + e^{\chi _4 } }}
\]
According to the network adversarial framework, the partial derivative of ${\rm{Tanh}}(\chi _1 ,\chi _2 ,\chi _3 ,\chi _4 )$ can be expressed as 
\begin{equation}\label{f10}
\setlength{\arraycolsep}{2pt} 
\renewcommand{\arraystretch}{2} 
\left\{ \begin{array}{l}
 \frac{{\partial {\rm{Tanh}}(\chi _1 ,\chi _2 ,\chi _3 ,\chi _4 )}}{{\partial \chi _1 }} = \frac{{e^{\chi _1 } }}{{e^{\chi _3 }  + e^{ - \chi _4 } }} \\ 
 \frac{{\partial {\rm{Tanh}}(\chi _1 ,\chi _2 ,\chi _3 ,\chi _4 )}}{{\partial \chi _2 }} = \frac{{e^{ - \chi _2 } }}{{e^{\chi _3 }  + e^{ - \chi _4 } }} \\ 
 \frac{{\partial {\rm{Tanh}}(\chi _1 ,\chi _2 ,\chi _3 ,\chi _4 )}}{{\partial \chi _3 }} = \frac{{\left( {e^{ - \chi _2 }  - e^{\chi _1 } } \right)e^{\chi _3 } }}{{\left( {e^{\chi _3 }  + e^{ - \chi _4 } } \right)^2 }} \\ 
 \frac{{\partial {\rm{Tanh}}(\chi _1 ,\chi _2 ,\chi _3 ,\chi _4 )}}{{\partial \chi _4 }} = \frac{{\left( {e^{\chi _1 }  - e^{ - \chi _2 } } \right)e^{ - \chi _4 } }}{{\left( {e^{\chi _3 }  + e^{ - \chi _4 } } \right)^2 }} \\ 
 \end{array} \right.
\end{equation}
Taking the reciprocal of each partial derivative in formula \ref{f10} is the partial derivative of the adversarial function $\xi _{{\rm{Tanh}}} (\chi _1 ,\chi _2 ,\chi _3 ,\chi _4 )$, which can be expressed as follows

\begin{equation}\label{f11}
\setlength{\arraycolsep}{2pt} 
\renewcommand{\arraystretch}{2} 
\left\{ \begin{array}{l}
 \frac{{\partial \xi _{{\rm{Tanh}}} (\chi _1 ,\chi _2 ,\chi _3 ,\chi _4 )}}{{\partial \chi _1 }} = \frac{{e^{\chi _3 }  + e^{ - \chi _4 } }}{{e^{\chi _1 } }} \\ 
 \frac{{\partial \xi _{{\rm{Tanh}}} (\chi _1 ,\chi _2 ,\chi _3 ,\chi _4 )}}{{\partial \chi _2 }} = \frac{{e^{\chi _3 }  + e^{ - \chi _4 } }}{{e^{ - \chi _2 } }} \\ 
 \frac{{\partial \xi _{{\rm{Tanh}}} (\chi _1 ,\chi _2 ,\chi _3 ,\chi _4 )}}{{\partial \chi _3 }} = \frac{{\left( {e^{\chi _3 }  + e^{ - \chi _4 } } \right)^2 }}{{\left( {e^{ - \chi _2 }  - e^{\chi _1 } } \right)e^{\chi _3 } }} \\ 
 \frac{{\partial \xi _{{\rm{Tanh}}} (\chi _1 ,\chi _2 ,\chi _3 ,\chi _4 )}}{{\partial \chi _4 }} = \frac{{\left( {e^{\chi _3 }  + e^{ - \chi _4 } } \right)^2 }}{{\left( {e^{\chi _2 }  - e^{ - \chi _2 } } \right)e^{ - \chi _4 } }} \\ 
 \end{array} \right.
\end{equation}
Set the constant coefficients in each partial derivative as trainable parameter $\alpha_i$, initialize $\alpha_i$ as a unit constant according to the formula (Initialization can be found in Appendix \ref{A-2-1}). Clculate the partial derivatives and integrals in formula (\ref{f11}) according to the operation rules when splitting, the adversarial function of ${\rm{Tanh}}(\chi _1 ,\chi _2 ,\chi _3 ,\chi _4 )$ is obtained as follows 

\begin{equation}\label{f12}
\setlength{\arraycolsep}{2pt} 
\renewcommand{\arraystretch}{2} 
\begin{array}{l}
  - \left( {\alpha _1  + \alpha _2  + \alpha _4  + \alpha _5 } \right){\rm{e}}^{ - \chi _1 }  + \frac{{\alpha _7 \left( {{\rm{e}}^{ - \chi _3 } \alpha _8 ^{ - 2}  - {\rm{e}}^{ - \chi _3 }  - 2\alpha _8 ^{ - 1} \chi _3 } \right)}}{{\alpha _7 \alpha _8  - 1}} +  \\ 
 \frac{{\alpha _{10} \left( {{\rm{e}}^{\chi _4 } \alpha _2 ^2  - {\rm{e}}^{ - \chi _4 }  + 2\alpha _{12} \chi _4 } \right)}}{{\alpha _{10} \alpha _{11}  - 1}} + \alpha _3  + \alpha _6  + \alpha _9  + \alpha _{12}  \\ 
 \end{array}
\end{equation}

\section{Experiment} \label{5}
In this section, we presented our experimental results on image classification using the CIFAR10 \cite{sharma2018analysis} and CIFAR100 \cite{sharma2018analysis} datasets, demonstrating the effectiveness of our network adversarial method in improving model performance.The specific structures of the VIT, ResNet, and SwT models in this experiment are VIT Base, ResNet34, and Swin Transformer tiny, respectively. We implemented adversarization based on Sigmoid, Tanh, and GeLu activation functions using PyTorch 2.1.2. Additionally, we employed various HD-FGD for Tanh and GeLu. It is worth noting that although the decomposition process for GeLu is more complex, the resulting adversarial function is relatively concise (reference to Appendix \ref{A-2-2}). All experiments were conducted on NVIDIA GeForce RTX 3090 with 24GB memory, and the reported results are averages over 5 random seeds. Furthermore, we applied our method to machine translation tasks in the NLP domain (reference to Appendix \ref{A-4}), where we observed significant improvements in model performance.

\subsection{GA is Effective}
The simplified form of activation functions only requires the GA to achieve gradient correction during training. Taking ${\rm{Sigmoid}}_\vartheta  (x)$ in \ref{4-1} as an example, we compared the network architecture performance of VIT, ResNet, and Swin-Transformer (SwT) models on different image classification tasks by alternately using the activation function as either $\xi _\vartheta  (x)$ or ${\rm{Sigmoid}}_\vartheta  (x)$ in the linear and convolutional layers. Table \ref{table-1} shows that all models consistently outperformed the baseline in terms of GA ACC and Top-5 Error, providing 3.87\% to 1466.11\% ACC and 7.00\% to 82.00\% Top-5 Error performance improvements for different models, respectively. It can be clearly seen from Figure \ref{fig:4} that the loss curve is significantly more stable after using the GA in different task scenarios.

\subsection{HD-FGD is Effective} \label{5-2}
It's worth noting that the HD-FGD not only provides generality for network adversarial training but can also be individually applied to ordinary networks to replace the original activation functions. Tanh and Gelu are both decomposed into four components using the HD-FGD method, and the decomposed activation functions are used in place of the original activation functions during model training in the linear layers. Table \ref{table-2} shows the new activation functions obtained by various models through the HD-FGD method, achieving higher ACC and Top5-Error on different datasets compared to the original activation functions. For more detailed experimental analysis, please refer to Appendix \ref{A-3-2}. Since the HD-FGD uses parallel computation to solve the parameter values of trainable matrices, it can improve the training speed of all models. Furthermore, in \ref{5-3}, it is entirely feasible to replace the original activation functions with the decomposed activation functions obtained through HD-FGD.

\subsection{SA is Effective} \label{5-3}
To further evaluate the effectiveness of SA in network adversarial methods, we based our analysis on the experiments outlined in Section \ref{5-2}. We utilized the activation functions optimized by HD-FGD for Tanh and GeLu as the original activation functions for each model. Subsequently, SA was employed for model training. The results presented in Table \ref{table-3} indicate that applying SA on top of HD-FGD significantly enhances model prediction performance, with a maximum improvement of 104.64\% in ACC. For a more detailed analysis of the experiments, please refer to Appendix \ref{A-3-3}. The effectiveness of SA primarily depends on the complexity of the activation function decomposition. If HD-FGD proves effective, further leveraging SA will become the preferred method for enhancing model performance.

\section{Conclusion}
In this work, we propose a network adversarial method that achieves gradient adversarial training during model training. We demonstrate the experimental results of using this method with various activation functions compared to the current conventions in different domains. The results show that both GA and SA significantly improve model prediction performance. Additionally, HD-FGD accelerates model training speed and contributes to notable performance improvements. Based on our method, we recommend researchers who still use standard activation functions to rewrite their networks at a lower cost. HD-FGD and network adversarial method can potentially bring substantial performance enhancements to the rewritten models.

The network adversarial method innovatively uses a combination of two functions during training as the activation of network layers, alternately correcting the gradients. In the future, it is possible to consider using functions with different properties or more function combinations as the nonlinear activations of the network to achieve smooth adversarialization between gradients.

The introduction of HD-FGD splits the original activation function into multiple terms while also transforming the activation function from a two-dimensional space to a higher-dimensional space. The projection of the spatial activation function onto various planes reflects the mapping of the derivatives of different decomposition terms in the two-dimensional plane, providing feasibility and new ideas for exploring the activation function in higher dimensions.

\textbf{Limitations} \quad Although the network advisory method has achieved exciting results, there are also some remaining issues. Due to the need to customize the number of items and expression of the original activation function for HD-FGD, different definitions will bring different gain effects. In addition, the solution of the adversarial function depends on the specific expression of each splitting term. If the final effect does not reach the ideal state, it is necessary to redefine the splitting form or number of splitting terms of the activation function from the HD-FGD stage. However, as shown in our experiment, our method remains effective even though custom split forms may bring different effects. Although it is necessary to manually derive integrals during the solving of adversarial functions, the solved adversarial functions can be applied to all models.

\bibliographystyle{plain}
\bibliography{ref.bib}

\newpage
\appendix

\section{Appendix}
\subsection{Formula Derivation} 
\subsubsection{Solving the Adversarial Function of Sigmoid} \label{A-1-1}
By using the method of substitution to solve this integral, let's define $u = e^x $. Then ${\rm{d}}u = e^x {\rm{d}}x$, the integral of $\xi _{{\rm{Sigmoid}}} (x)^{'}$ can be computed as follows
\[
\setlength{\arraycolsep}{2pt} 
\renewcommand{\arraystretch}{2} 
\begin{array}{l}
 \int {\frac{{\left( {{\rm{e}}^x  + 1} \right)^2 }}{{{\rm{e}}^x }}} {\rm{d}}x \\ 
  = \int {\frac{2}{u}}  + \frac{1}{{u^2 }} + 1{\rm{d}}u \\ 
  = 2\int {\frac{1}{u}} {\rm{d}}u + \int {\frac{1}{{u^2 }}} {\rm{d}}u + \int {1{\rm{d}}u}  \\ 
  = 2\ln \left( {\left| u \right|} \right) + \frac{{u^2  - 1}}{u} \\ 
  = \frac{{{\rm{e}}^{2x}  - 1}}{{{\rm{e}}^x }} + 2x + C \\ 
 \end{array}
\]

\subsubsection{Improved Sigmoid Adversarial Function Solution} \label{A-1-2}
By using the method of substitution to solve this integral and defining $u = e^x$, we obtain ${\rm{d}}u = e^x {\rm{d}}x$. The integral of $\xi _{{\rm{Sigmoid}}_\vartheta  } (x)'$ can be computed as follows

\begin{equation}\label{f13}
\setlength{\arraycolsep}{2pt} 
\renewcommand{\arraystretch}{2} 
\begin{array}{l}
 \int {\frac{{\left( {{\rm{e}}^x  + 1} \right)^2 }}{{{\rm{e}}^x  + \alpha \left( {{\rm{e}}^x  + 1} \right)^2 }}} {\rm{d}}x \\ 
  = \int {\frac{{u^2  + 2u + 1}}{{\alpha u^3  + \left( {2\alpha  + 1} \right)u^2  + \alpha u}}{\rm{d}}} u \\ 
  = \int {\frac{{u^2  + 2u + 1}}{{u\left( {\alpha u^2  + \left( {2\alpha  + 1} \right)u + \alpha } \right)}}{\rm{d}}} u \\ 
  = \int {\frac{{u^2  + 2u + 1}}{{\alpha u\left( {u - \frac{{\sqrt {4\alpha  + 1}  - 2\alpha  - 1}}{{2\alpha }}} \right)\left( {u + \frac{{\sqrt {4\alpha  + 1}  + 2\alpha  + 1}}{{2\alpha }}} \right)}}} {\rm{d}}u \\ 
 \end{array}
\end{equation}
Partial fraction decomposition of formula (\ref{f13})
\begin{equation}\label{f14}
\setlength{\arraycolsep}{2pt} 
\renewcommand{\arraystretch}{2} 
\begin{array}{l}
 \int {\frac{{u^2  + 2u + 1}}{{\alpha u\left( {u - \frac{{\sqrt {4\alpha  + 1}  - 2\alpha  - 1}}{{2\alpha }}} \right)\left( {u + \frac{{\sqrt {4\alpha  + 1}  + 2\alpha  + 1}}{{2\alpha }}} \right)}}{\rm{d}}u}  \\ 
  = \int {\frac{1}{\alpha } \cdot \left( {\frac{A}{{u + \frac{{\sqrt {4\alpha  + 1}  + 2\alpha  + 1}}{{2\alpha }}}} + \frac{B}{{u - \frac{{\sqrt {4\alpha  + 1}  - 2\alpha  - 1}}{{2\alpha }}}} + \frac{C}{u}} \right)} {\rm{d}}u \\ 
  = \int {\frac{1}{\alpha } \cdot \frac{{\left( {2\alpha C + 2\alpha B + 2\alpha A} \right)u^2  + \left( {\left( {4\alpha  + 2} \right)C + \left( {2\alpha  + 1} \right)B + \left( {2\alpha  + 1} \right)A} \right)u + \sqrt {4\alpha  + 1} \left( {B - A} \right)u + 2\alpha C}}{{2u\left( {u - \frac{{\sqrt {4\alpha  + 1}  - 2\alpha  - 1}}{{2\alpha }}} \right)\left( {u + \frac{{\sqrt {4\alpha  + 1}  + 2\alpha  + 1}}{{2\alpha }}} \right)}}} {\rm{d}}u \\ 
 \end{array}
\end{equation}
From formula (\ref{f14}), we can obtain a system of equations based on the coefficients of equal powers
\begin{equation}\label{f15}
\setlength{\arraycolsep}{2pt} 
\renewcommand{\arraystretch}{2} 
\left\{ \begin{array}{l}
 u^0 :C = 1 \\ 
 u^1 :\frac{C}{\alpha } + 2C + \frac{{B\sqrt {4\alpha  + 1} }}{{2\alpha }} + \frac{B}{{2\alpha }} + B - \frac{{A\sqrt {4\alpha  + 1} }}{{2\alpha }} + \frac{A}{{2\alpha }} + A = 2 \\ 
 u^2 :A + B + C = 1 \\ 
 \end{array} \right. \Rightarrow \left\{ \begin{array}{l}
 A = \frac{1}{{\sqrt {4\alpha  + 1} }} \\ 
 B =  - \frac{1}{{\sqrt {4\alpha  + 1} }} \\ 
 \end{array} \right.
\end{equation}
Substituting the result of formula (\ref{f15}) back into formula (\ref{f14}) yields $\xi _{{\rm{Sigmoid}}_\vartheta  } (x)$
\begin{equation}\label{f16}
\setlength{\arraycolsep}{2pt} 
\renewcommand{\arraystretch}{2.5} 
\begin{array}{l}
 \frac{1}{\alpha }\int {\frac{1}{{\sqrt {4\alpha  + 1} \left( {u + \frac{{\sqrt {4\alpha  + 1}  + 2\alpha  + 1}}{{2\alpha }}} \right)}}}  - \frac{1}{{\sqrt {4\alpha  + 1} \left( {u + \frac{{\sqrt {4\alpha  + 1}  - 2\alpha  - 1}}{{2\alpha }}} \right)}} + \frac{1}{u}{\rm{d}}u \\ 
  = \frac{1}{\alpha }\left( {\frac{1}{{\sqrt {4\alpha  + 1} }}\underbrace {\int {\frac{1}{{u + \frac{{\sqrt {4\alpha  + 1}  + 2\alpha  + 1}}{{2\alpha }}}}} {\rm{d}}u}_{(1)} - \frac{1}{{\sqrt {4\alpha  + 1} }}\underbrace {\int {\frac{1}{{u - \frac{{\sqrt {4\alpha  + 1}  - 2\alpha  - 1}}{{2\alpha }}}}} {\rm{d}}u}_{(2)} + \int {\frac{1}{u}{\rm{d}}u} } \right) \\ 
  = \frac{{{\rm{ln}}\left( {\left| {2\alpha u + \sqrt {4\alpha  + 1}  + 2\alpha  + 1} \right|} \right)}}{{\alpha \sqrt {4\alpha  + 1} }} - \frac{{{\rm{ln}}\left( {\left| {2\alpha u - \sqrt {4\alpha  + 1}  + 2\alpha  + 1} \right|} \right)}}{{\alpha \sqrt {4\alpha  + 1} }} + \frac{{\ln \left( {\left| u \right|} \right)}}{\alpha } \\ 
  =  - \frac{{\ln \left( {|2\alpha {\rm{e}}^{\rm{x}}  - \sqrt {4\alpha  + 1}  + 2\alpha  + 1|} \right)}}{{\alpha \sqrt {4\alpha  + 1} }} + \frac{{\ln \left( {|2\alpha {\rm{e}}^{\rm{x}}  + \sqrt {4\alpha  + 1}  + 2\alpha  + 1|} \right)}}{{\alpha \sqrt {4\alpha  + 1} }} + \frac{x}{\alpha } + C \\ 
 \end{array}
\end{equation}
For the (1) part in formula (\ref{f16}), we will solve it using the method of substitution. Let's assume $v = 2\alpha u + \sqrt {4\alpha  + 1}  + 2\alpha  + 1$ as the substitution variable. Then $u = \frac{{v - \sqrt {4\alpha  + 1}  - 2\alpha  - 1}}{{2\alpha }}$ as ${\rm{d}}u = \frac{1}{{2\alpha }}{\rm{d}}v$. Further steps for the computation are as follow
\[
\setlength{\arraycolsep}{2pt} 
\renewcommand{\arraystretch}{2} 
\begin{array}{l}
 \int {\frac{1}{{u + \frac{{\sqrt {4\alpha  + 1}  + 2\alpha  + 1}}{{2\alpha }}}}} {\rm{d}}u \\ 
  = \int {\frac{{2\alpha }}{{2\alpha u + \sqrt {4\alpha  + 1}  + 2\alpha  + 1}}} {\rm{d}}u \\ 
  = 2\alpha \int {\frac{1}{{2\alpha v}}} {\rm{d}}v \\ 
  = \ln \left( {\left| v \right|} \right) \\ 
  = \ln \left( {\left| {2\alpha u + \sqrt {4\alpha  + 1}  + 2\alpha  + 1} \right|} \right) \\ 
 \end{array}
\]
For the (2) part in formula (\ref{f16}), we will solve it using the method of substitution. Let's assume $v = 2\alpha u - \sqrt {4\alpha  + 1}  + 2\alpha  + 1$ as the substitution variable. Then $u = \frac{{v + \sqrt {4\alpha  + 1}  - 2\alpha  - 1}}{{2\alpha }}$. Further, we have ${\rm{d}}u = \frac{1}{{2\alpha }}{\rm{d}}v$. The specific computation process is as follows
\[
\setlength{\arraycolsep}{2pt} 
\renewcommand{\arraystretch}{2} 
\begin{array}{l}
 \int {\frac{1}{{u - \frac{{\sqrt {4\alpha  + 1}  - 2\alpha  - 1}}{{2\alpha }}}}} {\rm{d}}u \\ 
  = \int {\frac{{2\alpha }}{{2\alpha u - \sqrt {4\alpha  + 1}  + 2\alpha  + 1}}} {\rm{d}}u \\ 
  = 2\alpha \int {\frac{1}{{2\alpha v}}} {\rm{d}}v \\ 
  = \ln \left( {\left| v \right|} \right) \\ 
  = \ln \left( {\left| {2\alpha u - \sqrt {4\alpha  + 1}  + 2\alpha  + 1} \right|} \right) \\ 
 \end{array}
\]

\subsection{Adversarial Functions of Other Activation Functions}
\subsubsection{Tanh Trainable Parameter Initialization} \label{A-2-1}
The partial derivative of $\xi _{{\rm{Tanh}}} (\chi _1 ,\chi _2 ,\chi _3 ,\chi _4 )$ can be obtained from formula (\ref{f11}), and the result of integrating each partial derivative is as follows
\begin{equation}\label{f17}
\setlength{\arraycolsep}{2pt} 
\renewcommand{\arraystretch}{2.5} 
\left\{ \begin{array}{l}
 \int {\frac{{\partial \xi _{{\rm{Tanh}}} (\chi _1 ,\chi _2 ,\chi _3 ,\chi _4 )}}{{\partial \chi _1 }}} {\rm{d}}\chi _1  = {\rm{ - }}\left( {{\rm{e}}^{\chi _3 }  + {\rm{e}}^{ - \chi _4 } } \right){\rm{e}}^{{\rm{ - }}\chi _{\rm{1}} }  + C \\ \hspace{3.6cm} =  - \left( {\alpha _1  + \alpha _2 } \right){\rm{e}}^{{\rm{ - }}\chi _{\rm{1}} }  + \alpha _3  \\ 
 \int {\frac{{\partial \xi _{{\rm{Tanh}}} (\chi _1 ,\chi _2 ,\chi _3 ,\chi _4 )}}{{\partial \chi _2 }}} {\rm{d}}\chi _2  = {\rm{ - }}\left( {{\rm{e}}^{\chi _3 }  + {\rm{e}}^{ - \chi _4 } } \right) \cdot {\rm{e}}^{{\rm{ - }}\chi _{\rm{2}} }  + C \\ \hspace{3.6cm} =  - \left( {\alpha _4  + \alpha _5 } \right){\rm{e}}^{{\rm{ - }}\chi _{\rm{1}} }  + \alpha _6  \\ 
 \int {\frac{{\partial \xi _{{\rm{Tanh}}} (\chi _1 ,\chi _2 ,\chi _3 ,\chi _4 )}}{{\partial \chi _3 }}} {\rm{d}}\chi _3  = \frac{{{\rm{e}}^{\chi _2 } \left( {{\rm{e}}^{ - \chi _3  - 2\chi _4 }  - {\rm{e}}^{\chi _3 }  - 2{\rm{e}}^{ - \chi _4 } \chi _3 } \right)}}{{{\rm{e}}^{\chi _1  + \chi _2 }  - 1}} + C \\ \hspace{3.6cm} = \frac{{\alpha _7 \left( {{\rm{e}}^{ - \chi _3 } \alpha _8 ^{ - 2}  - {\rm{e}}^{\chi _3 }  - 2\alpha _8 ^{ - 1} \chi _3 } \right)}}{{\alpha _7 \alpha _8  - 1}} + \alpha _9  \\ 
 \int {\frac{{\partial \xi _{{\rm{Tanh}}} (\chi _1 ,\chi _2 ,\chi _3 ,\chi _4 )}}{{\partial \chi _4 }}} {\rm{d}}\chi _4  = \frac{{{\rm{e}}^{\chi _2 } \left( {{\rm{e}}^{\chi _4  + 2\chi _3 }  - {\rm{e}}^{ - \chi _4 }  - 2{\rm{e}}^{\chi _3 } \chi _4 } \right)}}{{{\rm{e}}^{\chi _1  + \chi _2 }  - 1}} + C\\ \hspace{3.6cm} = \frac{{\alpha _{10} \left( {{\rm{e}}^{\chi _4 } \alpha _2 ^{\rm{2}}  - {\rm{e}}^{\chi _4 }  + 2\alpha _{12} \chi _4 } \right)}}{{\alpha _{10} \alpha _{11}  - 1}} + \alpha _{12}  \\ 
 \end{array} \right.
\end{equation}
$\alpha_i$ in formula (\ref{f17}) represents trainable parameters. The trainable parameter in formula (\ref{f17}) can be initialized to the 1 unit constant calculated by the formula

\[
\setlength{\arraycolsep}{2pt} 
\renewcommand{\arraystretch}{1.5} 
\left\{ \begin{array}{l}
 \alpha _1  = \alpha _4  = {\rm{e}}^{\chi _3 }  \Rightarrow {\rm{e}} \\ 
 \alpha _2  = \alpha _5  = {\rm{e}}^{ - \chi _4 }  \Rightarrow {\rm{e}} \\ 
 \alpha _3  = \alpha _6  = \alpha _9  = \alpha _{12}  = 0 \\ 
 \alpha _7  = \alpha _{10}  = {\rm{e}}^{\chi _2 }  \Rightarrow {\rm{e}} \\ 
 \alpha _8  = \alpha _{11}  = e \\ 
 \end{array} \right.
\]

\subsubsection{ The Adversarial Function of GeLu} \label{A-2-2}
\begin{figure}[h]
\centering
\begin{minipage}{0.48\textwidth}
  \centering
  \includegraphics[width=\textwidth]{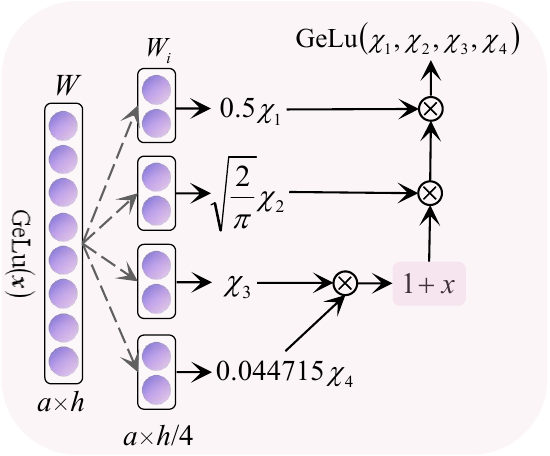}
  \caption{HD-FGD splits GeLu into 4 terms.}
  \label{fig:gleu}
\end{minipage}\hfill
\begin{minipage}{0.45\textwidth}
  \centering
  \includegraphics[width=\textwidth]{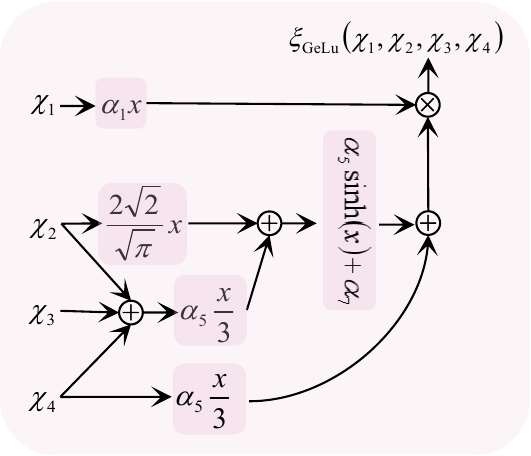}
  \caption{Calculation structure diagram of ${\xi _{{\rm{GeLu}}} (\chi _1 ,\chi _2 ,\chi _3 ,\chi _4 )}$.}
  \label{fig:gelu-com}
\end{minipage}
\end{figure}

GeLu is most widely used in models based on transformer architecture, with the expression $0.5x(1 + \rm{Tanh} (\sqrt {\frac{2}{\pi }} (x + 0.44715x^3 )))$, which is represented as formula (\ref{f18}) after passing through a linear layer
\begin{equation}\label{f18}
{\rm{GeLu}}(W,x) = 0.5Wx(1 + {\rm{Tanh}}(\sqrt {\frac{2}{\pi }} Wx(1 + Wx \cdot 0.44715Wx)))
\end{equation}

In the above formula, $W$ represents the linear layer parameter matrix. As shown in Figure \ref{fig:gleu}, splitting GeLu into 4 terms can be represented as
\begin{equation}\label{f19}
{\rm{GeLu}}(W_1 ,W_2 ,W_3 ,W_4 ,x) = 0.5W_1 x(1 + {\rm{Tanh}}(\sqrt {\frac{2}{{\rm{\pi }}}} W_2 x(1 + W_3 x \cdot 0.44715W_4 x)))
\end{equation}
Let $\chi _i  = W_i x$, then formula (\ref{f19}) can be modified to
\[
{\rm{GeLu}}(\chi _1 ,\chi _2 ,\chi _3 ,\chi _4 ) = 0.5\chi _1 (1 + {\rm{Tanh}}(\sqrt {\frac{2}{{\rm{\pi }}}} \chi _2 (1 + \chi _3  \cdot 0.44715\chi _4 )))
\]
According to the network adversarial method, the partial derivative of ${\rm{GeLu}}(\chi _1 ,\chi _2 ,\chi _3, \chi _4 )
$ can be solved as formula (\ref{f20})
\begin{equation}\label{f20}
\setlength{\arraycolsep}{2pt} 
\renewcommand{\arraystretch}{2.5} 
\left\{ \begin{array}{l}
 \frac{{\partial {\rm{GeLu}}(\chi _1 ,\chi _2 ,\chi _3 ,\chi _4 )}}{{\partial \chi _1 }} = 0.5\left( {{\rm{Tanh}}\left( {\sqrt {\frac{2}{\pi }\chi _2 } \left( {1 + \chi _3  \cdot 0.044715\chi _4 } \right)} \right)} \right) \\ 
 \frac{{\partial {\rm{GeLu}}(\chi _1 ,\chi _2 ,\chi _3 ,\chi _4 )}}{{\partial \chi _2 }} = \frac{{8943\sqrt 2 \chi _1 \chi _3 \chi _4  + 3125.64\sqrt 2 \chi _1 }}{{400000\sqrt \pi  \cosh ^2 \left( {\frac{{8943\chi _3 \chi _4 \chi _2 }}{{3125.32\sqrt {2\pi } }} + \sqrt {\frac{2}{\pi }} \chi _2 } \right)}} \\ 
 \frac{{\partial {\rm{GeLu}}(\chi _1 ,\chi _2 ,\chi _3 ,\chi _4 )}}{{\partial \chi _3 }} = \frac{{8943\chi _1 \chi _2 \chi _4 }}{{3125.64\sqrt {2\pi } \cosh ^2 \left( {\frac{{8943\chi _2 \chi _4 \chi _3 }}{{3125.32\sqrt {2\pi } }} + \sqrt {\frac{2}{\pi }} \chi _2 } \right)}} \\ 
 \frac{{\partial {\rm{GeLu}}(\chi _1 ,\chi _2 ,\chi _3 ,\chi _4 )}}{{\partial \chi _4 }} = \frac{{8943\chi _1 \chi _2 \chi _3 }}{{3125.64\sqrt {2\pi } \cosh ^2 \left( {\frac{{8943\chi _2 \chi _3 \chi _4 }}{{3125.32\sqrt {2\pi } }} + \sqrt {\frac{2}{\pi }} \chi _2 } \right)}} \\ 
 \end{array} \right.
\end{equation}
Based on this, the partial derivative of $\xi _{{\rm{GeLu}}} \left( {\chi _1 ,\chi _2 ,\chi _3 ,\chi _4 } \right)$ can be solved as the reciprocal of the partial derivatives of ${\rm{GeLu}}(\chi _1 ,\chi _2 ,\chi _3 ,\chi _4 )$ in formula (\ref{f20}), and the partial derivatives of $\xi _{{\rm{GeLu}}} \left( {\chi _1 ,\chi _2 ,\chi _3 ,\chi _4 } \right)$ can be integrated separately
\begin{equation}\label{f21}
\setlength{\arraycolsep}{2pt} 
\renewcommand{\arraystretch}{2.5} 
\left\{ \begin{array}{l}
 \int {\frac{{\partial \xi _{{\rm{GeLu}}} (\chi _1 ,\chi _2 ,\chi _3 ,\chi _4 )}}{{\chi _1 }}} {\rm{d}}\chi _1  = \frac{{\chi _1 }}{{0.5\left( {{\rm{1}} + {\rm{Tanh}}\left( {\sqrt {\frac{2}{\pi }} \chi _2 \left( {1 + \chi _3  \cdot 0.044715\chi _4 } \right)} \right)} \right)}} + C   = \alpha _1 \chi _1  + \alpha _2  \\ 
 \int {\frac{{\partial \xi _{{\rm{GeLu}}} (\chi _1 ,\chi _2 ,\chi _3 ,\chi _4 )}}{{\chi _2 }}} {\rm{d}}\chi _2  = \frac{{17757500000\pi  \cdot {\rm{sinh}}\left( {\frac{{{\rm{20325}}\sqrt {\rm{2}} \chi _3 \chi _4 \chi _2 }}{{7103\sqrt \pi  }} + 2\sqrt {\frac{2}{\pi }} \chi _2 } \right)}}{{4544161875\chi _1 \chi _3 ^2 \chi _4 ^2  + 4764322275\chi _1 \chi _3 \chi _4  + 1110071041\chi _1 }} +  \\ 
 \hspace{4cm} \frac{{78125.32\sqrt {{\rm{2}}\pi } \chi _2 }}{{223575\chi _1 \chi _3 \chi _4  + 78141\chi _1 }} + C \\  \hspace{3.6cm}= \alpha _3 \sinh \left( {\alpha _4 \beta _1 \beta _2 \chi _2  + \alpha _5 \chi _2 } \right) + \alpha _6 \chi _2  + \alpha _7  \\ 
 \int {\frac{{\partial \xi _{{\rm{GeLu}}} (\chi _1 ,\chi _2 ,\chi _3 ,\chi _4 )}}{{\chi _3 }}} {\rm{d}}\chi _3  = \frac{{185011841\pi  \cdot {\rm{sinh}}\left( {\frac{{{\rm{20325}}\sqrt {\rm{2}} \chi _2 \chi _4 \chi _3 }}{{7103\sqrt \pi  }} + 2\sqrt {\frac{2}{\pi }} \chi _2 } \right)}}{{3029441250\chi _1 \chi _2 ^2 \chi _4 ^2 }}  + \\ \hspace{4cm} \frac{{26047\sqrt \pi  \chi _3 }}{{74525\sqrt {\rm{2}} \chi _1 \chi _2 \chi _4 }} + C \\
 \hspace{3.6cm} = \alpha _8 \sinh \left( {\alpha _4 \beta _3 \beta _2 \chi _2  + \alpha _5 \chi _2 } \right) + \alpha _9 \chi _3  + \alpha _{10}  \\ 
 \int {\frac{{\partial \xi _{{\rm{GeLu}}} (\chi _1 ,\chi _2 ,\chi _3 ,\chi _4 )}}{{\chi _4 }}} {\rm{d}}\chi _4  = \frac{{185011841\pi  \cdot {\rm{sinh}}\left( {\frac{{{\rm{20325}}\sqrt {\rm{2}} \chi _2 \chi _3 \chi _4 }}{{7103\sqrt \pi  }} + 2\sqrt {\frac{2}{\pi }} \chi _2 } \right)}}{{3029441250\chi _1 \chi _2 ^2 \chi _3 ^2 }} + \\ \hspace{4cm} \frac{{26047\sqrt \pi  \chi _4 }}{{74525\sqrt {\rm{2}} \chi _1 \chi _2 \chi _3 }} + C \\
  \hspace{3.6cm}= \alpha _{11} \sinh \left( {\alpha _4 \beta _3 \beta _1 \chi _2  + \alpha _5 \chi _2 } \right) + \alpha _{12} \chi _4  + \alpha _{13}  \\ 
 \end{array} \right.
\end{equation}

$\alpha_i$ and $\beta_i$ in formula (\ref{f21}) both represent trainable parameters. 
$\alpha_2$, $\alpha_7$, $\alpha_{10}$ and $\alpha_{13}$ are the constant terms obtained after integration, and they are set as trainable parameters initialized to 0. Other trainable parameters can be initialized to the constant of 1 unit calculated from the equation, as shown in formula (\ref{22})

\begin{equation}\label{22}
\setlength{\arraycolsep}{2pt} 
\renewcommand{\arraystretch}{2} 
\left\{ \begin{array}{l}
 \alpha _1  = \frac{1}{{0.5\left( {{\rm{Tanh}}\left( {\sqrt {\frac{2}{\pi }} \chi _2 \left( {1 + \chi _3  \cdot 0.044715\chi _4 } \right)} \right)} \right)}} \Rightarrow \frac{1}{{0.5\left( {{\rm{Tanh}}\left( {\sqrt {\frac{2}{\pi }} \left( {1 + 0.044715} \right)} \right)} \right)}} \\ 
 \alpha _3  = \frac{{17757500000\pi }}{{4544161875\chi _1 \chi _3 ^2 \chi _4 ^2  + 4764322275\chi _1 \chi _3 \chi _4  + 1110071041\chi _1 }} \Rightarrow \frac{{17757500000\pi }}{{10418555191}} \\ 
 \alpha _4  = \frac{{20325\sqrt 2 }}{{7103\sqrt \pi  }} \\ 
 \beta _1  = \chi _3  \Rightarrow 1 \\ 
 \beta _2  = \chi _4  \Rightarrow 1 \\ 
 \beta _3  = \chi _2  \Rightarrow 1 \\ 
 \alpha _5  = 2\sqrt {\frac{2}{\pi }}  \\ 
 \alpha _6  = \frac{{78125.32\sqrt {2\pi } \chi _2 }}{{223575\chi _1 \chi _3 \chi _4  + 78141\chi _1 }} \Rightarrow \frac{{78125.32\sqrt {2\pi } }}{{301716}} \\ 
 \alpha _8  = \frac{{185011841\pi }}{{3029441250\chi _1 \chi _2 ^2 \chi _4 ^2 }} \Rightarrow \frac{{185011841\pi }}{{3029441250}} \\ 
 \alpha _9  = \frac{{26047\sqrt \pi  \chi _2 }}{{74525\sqrt 2 \chi _1 \chi _2 \chi _4 }} \Rightarrow \frac{{26047\sqrt \pi  }}{{74525\sqrt 2 }} \\ 
 \alpha _{11}  = \frac{{185011841\pi }}{{3029441250\chi _1 \chi _2 ^2 \chi _3 ^2 }} \Rightarrow \frac{{185011841\pi }}{{3029441250}} \\ 
 \alpha _{12}  = \frac{{26047\sqrt \pi  \chi _4 }}{{74525\sqrt 2 \chi _1 \chi _2 \chi _3 }} \Rightarrow \frac{{26047\sqrt \pi  }}{{74525\sqrt 2 }} \\ 
 \end{array} \right.
\end{equation}
$\xi _{{\rm{GeLu}}} (\chi _1 ,\chi _2 ,\chi _3, \chi _4 )$ can be represented as
\[
\begin{array}{l}
 \xi _{{\rm{GeLu}}} \left( {\chi _1 ,\chi _2 ,\chi _3 ,\chi _4 } \right) =\left({\alpha _1 \chi _1  + \alpha _2 } \right) \cdot \left( {\alpha _3 {\rm{sinh}}\left( {\alpha _4 \beta _1 \beta _2 \chi _2  + \alpha _5 \chi _2 } \right) + \alpha _6 \chi _2  + \alpha _7 } \right) \cdot  \\ 
 \begin{array}{*{20}c}
   {\begin{array}{*{20}c}
   {\begin{array}{*{20}c}
   {} & {}  \\
\end{array}} & {} & {} & {} \\
\end{array}} & {} & {} & {}  \hspace{0.5mm} \\
\end{array}\left( {\alpha _8 {\rm{sinh}}\left( {\alpha _4 \beta _2 \beta _3 \chi _3  + \alpha _5 \chi _2 } \right) + \alpha _9 \chi _3  + \alpha _{10} } \right) \cdot  \\ 
 \begin{array}{*{20}c}
   {\begin{array}{*{20}c}
   {\begin{array}{*{20}c}
   {} & {}  \\
\end{array}} & {} & {} & {} \\
\end{array}} & {} & {} & {} \hspace{0.5mm} \\
\end{array}\left( {\alpha _{11} {\rm{sinh}}\left( {\alpha _4 \beta _1 \beta _3 \chi _4  + \alpha _5 \chi _2 } \right) + \alpha _{12} \chi _4  + \alpha _{13} } \right) \\ 
 \end{array}
\]

In the splitting of formula (\ref{f21}), considering the complexity of the function and the possibility of overfitting with too many learnable parameters. As shown in Figure \ref{fig:gelu-com}, we tend to write the general formula of the function, simplify some learnable parameters, and reinitialize them as
\[
\setlength{\arraycolsep}{2pt} 
\renewcommand{\arraystretch}{2} 
\left\{ \begin{array}{l}
 \delta _1  = \frac{{\left( {\alpha _3  + \alpha _8  + \alpha _{11} } \right)}}{3} \\ 
 \delta _2  = \frac{{\left( {\alpha _6  + \alpha _9  + \alpha _{12} } \right)}}{3} \\ 
 \delta _3  = \frac{{\left( {\alpha _7  + \alpha _{10}  + \alpha _{13} } \right)}}{3} \\ 
 \beta = \beta _1  \beta _2 \beta _3\\
 \chi = \frac{{\left( {\chi _2  + \chi _3  + \chi _{4} } 
 \right)}}{3}\\
 \end{array} \right.
\]

$\xi _{{\rm{GeLu}}} \left({\chi _1 ,\chi _2 ,\chi _3, \chi _4} \right)$ can be rephrased as $\xi _{{\rm{GeLu}}} (\chi _1 ,\chi)$.
\[
\xi _{{\rm{GeLu}}} \left( {\chi _1 ,\chi} \right) = \left( {\alpha _1 \chi _1  + \alpha _2 } \right) \cdot \left( {\delta _1 {\rm{sinh}}\left( {\alpha _4 \beta \chi  + \alpha _5 \beta _2 } \right) + \delta _2 \chi  + \delta _3 } \right)
\]

\subsection{Analysis of Experimental Results}
\subsubsection{Analysis of GA Experiment Results} \label{A-3-1}
On the CIFAR10 dataset, using Sigmoid and its overall adversarial counterpart as the activation functions for the network layers of the VIT, SwT, and ResNet models resulted in improved prediction performance to varying degrees. Among them, ResNet showed the most significant improvement, with an increase of 101.52\% in ACC and a decrease of 82.00\% in Top-5 Error. In the CIFAR100 dataset, when Sigmoid was used as the activation function for the ResNet model in the classification task, it was almost ineffective in ResNet34. However, after applying the GA based on Sigmoid, there was a remarkable improvement in ACC with an increase of 1466.11\%, and a considerable decrease in Top-5 Error by 73.97\%.

\begin{table} 
\centering
\caption{ACC and Top-5 Error based on Sigmoid GA. GA represents global adversarial.}
\begin{tabular}{ccccc} 
\toprule
\textbf{Dataset} & \textbf{Model} & \textbf{GA} & \textbf{ACC} & \textbf{Top-5 Error} \\
\midrule
\multirow{6}{*}{\raisebox{-1.5ex}[0pt][0pt]{\rotatebox[origin=c]{90}{\textbf{CIFAR10}}}} & \multirow{2}{*}{VIT} & \checkmark & \textbf{63.26 $\pm$ 0.15}  & \textbf{3.12 $\pm$ 0.12}  \\
& & $\times$ & 55.17 $\pm$ 0.00 & 4.84  $\pm$ 0.00 \\
\cmidrule(lr){2-2}\cmidrule(lr){3-3}\cmidrule(lr){4-4}\cmidrule(lr){5-5}
& \multirow{2}{*}{ResNet} & \checkmark & \textbf{75.53 $\pm$ 0.32} & \textbf{1.55 $\pm$ 0.73}  \\
& & $\times$ & 37.48 $\pm$ 2.21 & 8.61 $\pm$ 0.07 \\
\cmidrule(lr){2-2}\cmidrule(lr){3-3}\cmidrule(lr){4-4}\cmidrule(lr){5-5}
& \multirow{2}{*}{SwT} & \checkmark & \textbf{50.14 $\pm$ 0.29}  & \textbf{6.91 $\pm$ 0.23}  \\
& & $\times$ & 48.27 $\pm$ 1.09 & 7.43 $\pm$ 0.45 \\
\midrule
\multirow{6}{*}{\raisebox{-1.5ex}[0pt][0pt]{\rotatebox[origin=c]{90}{\textbf{CIFAR100}}}} & \multirow{2}{*}{VIT} & \checkmark & \textbf{39.41 $\pm$ 0.00}  & \textbf{29.86 $\pm$ 0.00}  \\
& & $\times$ & 29.66 $\pm$ 0.00 & 39.95 $\pm$ 0.00 \\
\cmidrule(lr){2-2}\cmidrule(lr){3-3}\cmidrule(lr){4-4}\cmidrule(lr){5-5}
& \multirow{2}{*}{ResNet} & \checkmark & \textbf{47.14 $\pm$ 0.56}  & \textbf{22.76 $\pm$ 0.32}  \\
& & $\times$ & 3.01 $\pm$ 0.18 & 87.44 $\pm$ 0.65 \\
\cmidrule(lr){2-2}\cmidrule(lr){3-3}\cmidrule(lr){4-4}\cmidrule(lr){5-5}
& \multirow{2}{*}{SwT} & \checkmark & \textbf{24.41 $\pm$ 0.45}  & \textbf{47.76 $\pm$ 0.40}  \\
& & $\times$ & 19.56 $\pm$ 0.29 & 53.71 $\pm$ 0.45 \\
\bottomrule
\end{tabular}
\label{table-1}
\end{table}

\begin{figure}[h]
  \centering
  \begin{minipage}[b]{0.32\textwidth}
    \centering
    \includegraphics[width=\textwidth]{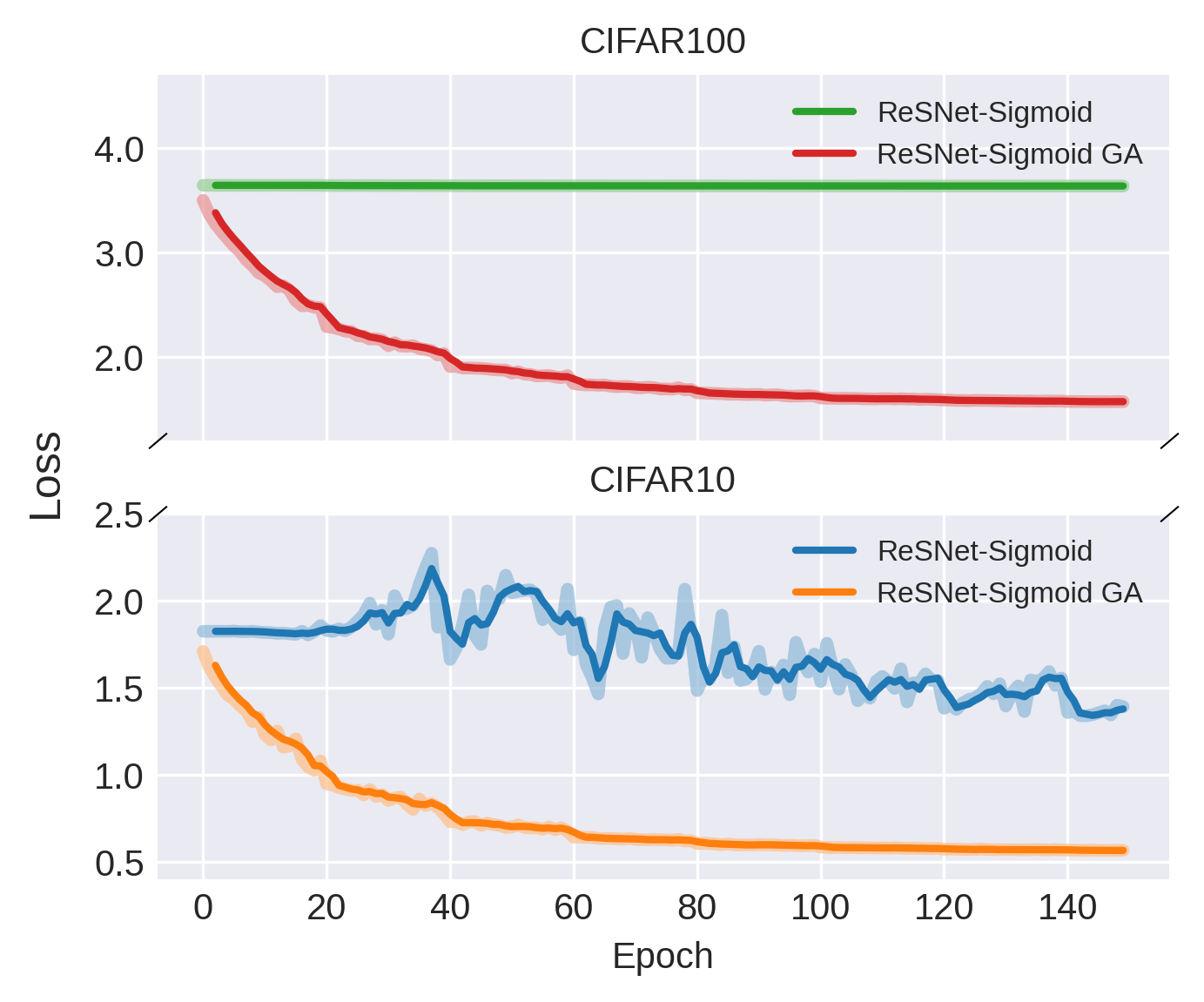}
    \subcaption{ResNet + Sigmoid}
  \end{minipage}
  \hfill
  \begin{minipage}[b]{0.32\textwidth}
    \centering
    \includegraphics[width=\textwidth]{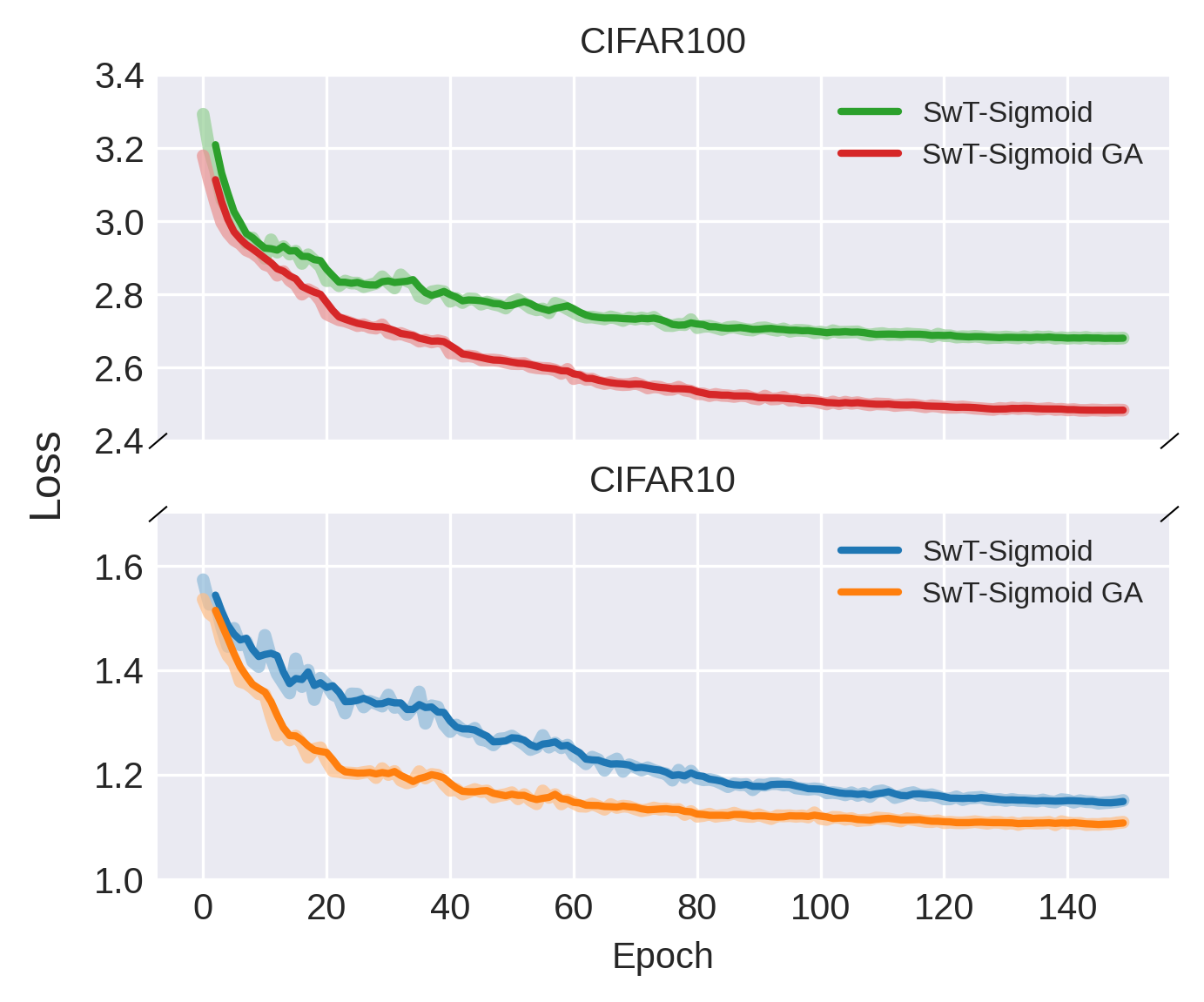}
    \subcaption{SwT + Sigmoid}
  \end{minipage}
  \hfill
  \begin{minipage}[b]{0.32\textwidth}
    \centering
    \includegraphics[width=\textwidth]{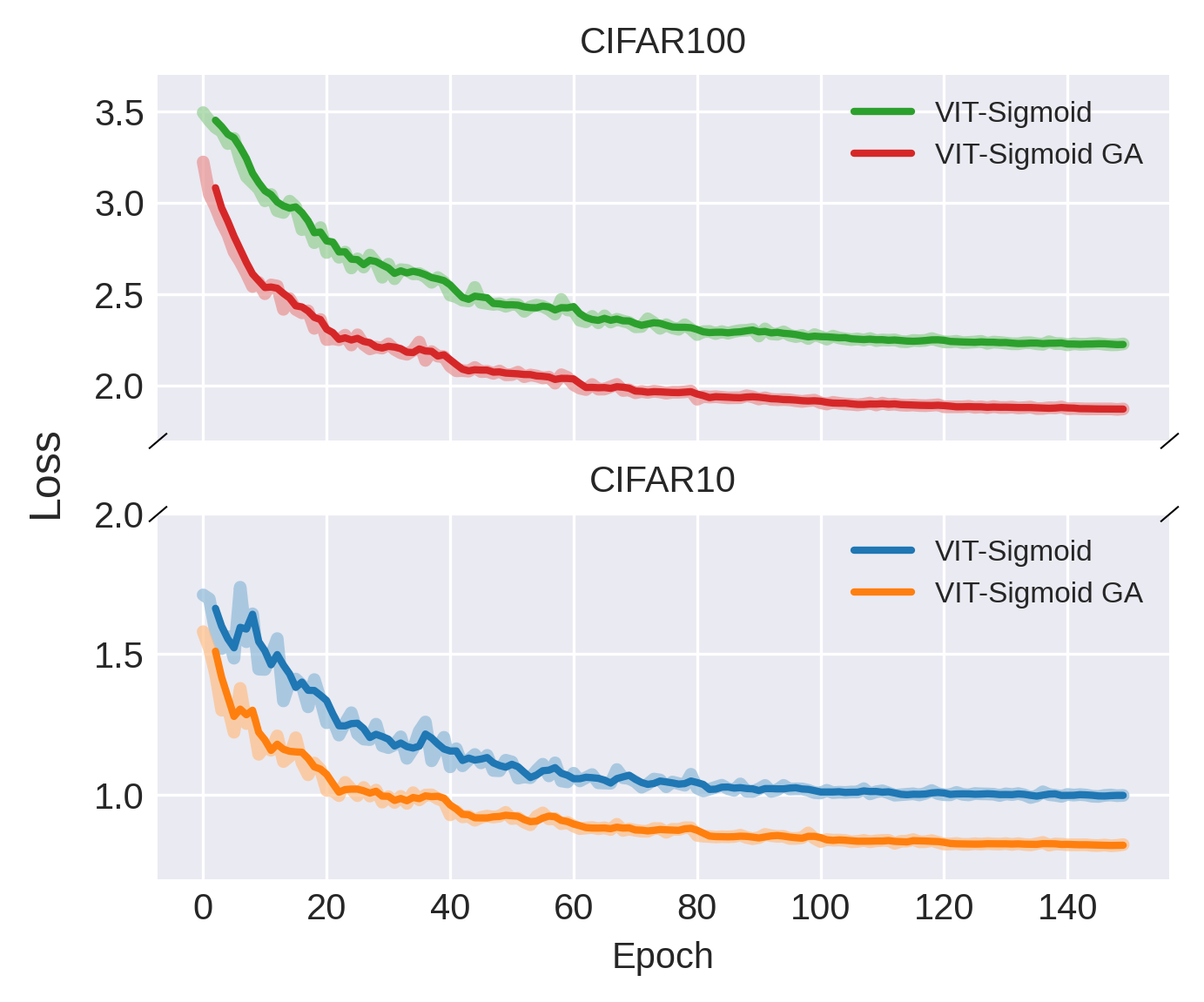}
    \subcaption{VIT + Sigmoid}
  \end{minipage}
  \caption{Compared with Sigmoid, the GA test losses of Sigmoid in different scenario tasks.}
  \label{fig:4}
\end{figure}

\subsubsection{Analysis of HD-FGD Experiment Results} \label{A-3-2}
In this experiment, Tanh and GeLu were used as activation function in the VIT and SwT models, respectively. The models were tested on the CIFAR10 and CIFAR100 datasets, and the results were used as the baseline. The HD-FGD was applied to split Tanh and GeLu into four components, and the decomposed forms were used to replace the original activation function. 

The results in Table \ref{table-2} showed that, for the CIFAR10 classification task, using HD-FGD resulted in ACC improvements ranging from 1.04\% to 68.89\%, Top-5 Error reductions ranging from 0.32\% to 73.22\%, and average runtime reductions per epoch ranging from 0.17\% to 10.41\%. For the CIFAR100 classification task, ACC improvements ranged from 3.77\% to 37.47\%, Top-5 Error reductions ranged from 2.10\% to 27.47\%, and average runtime reductions per epoch ranged from 2.17\% to 9.15\%.

\begin{figure}[h]
  \centering
  \begin{minipage}[b]{0.48\textwidth}
    \centering
    \includegraphics[width=\textwidth]{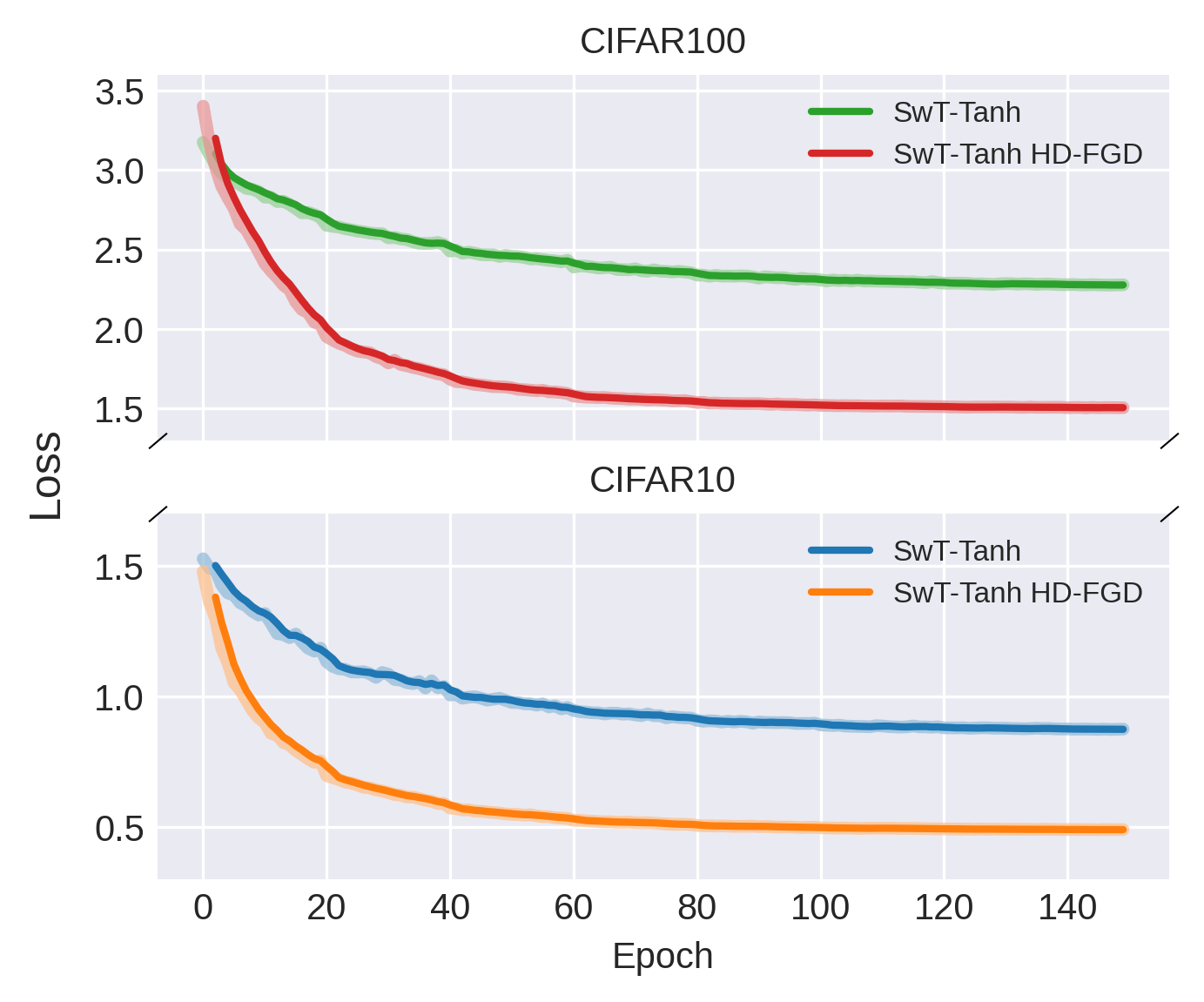}
    \subcaption{SwT + Tanh}
  \end{minipage}
  \hfill
  \begin{minipage}[b]{0.48\textwidth}
    \centering
    \includegraphics[width=\textwidth]{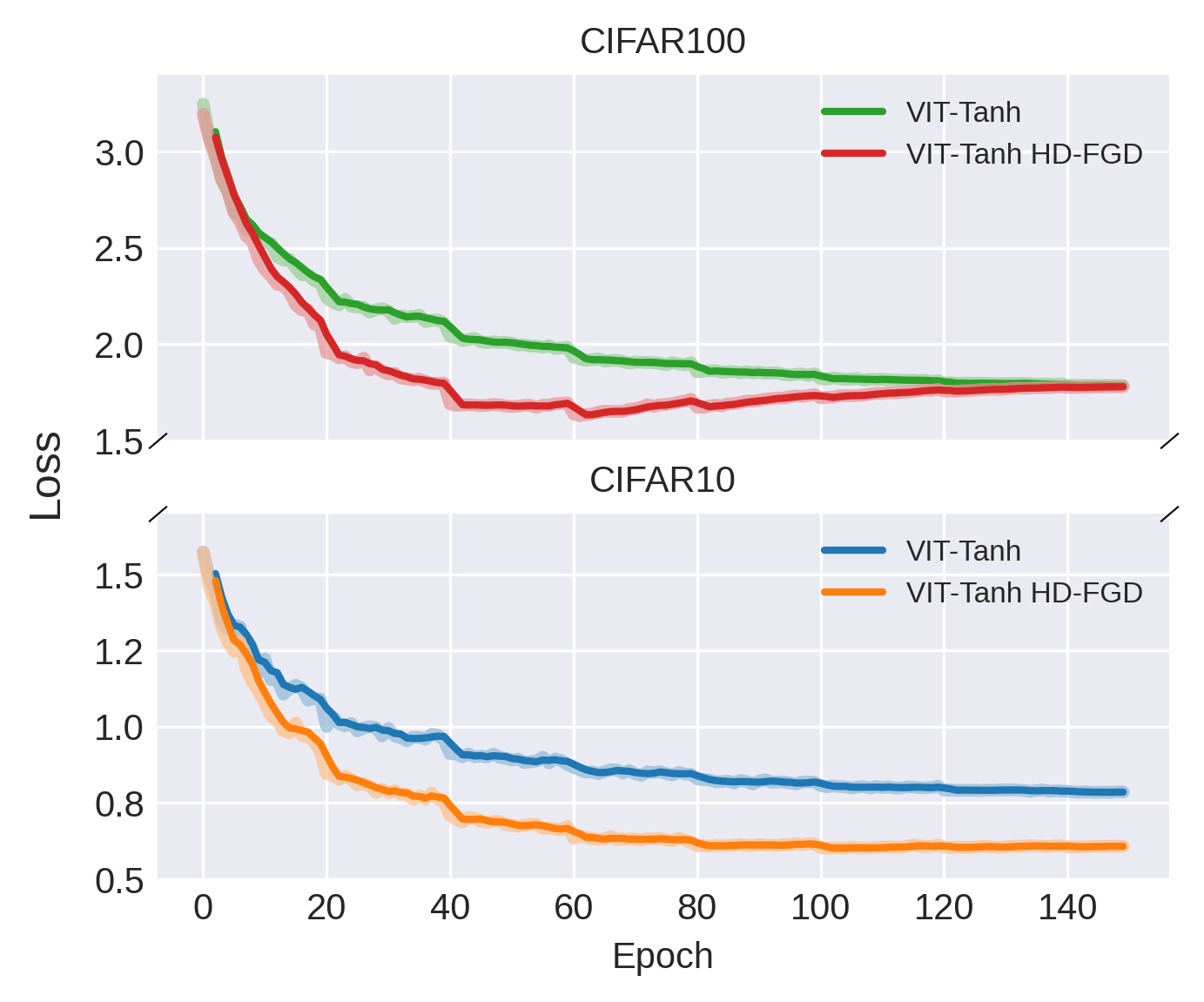}
    \subcaption{VIT + Tanh}
  \end{minipage}
  \caption{Compared with Tanh, HD-FGD test losses of Tanh in different scenario tasks.}
  \label{fig:5}
\end{figure}

\begin{figure}[h]
  \centering
  \begin{minipage}[b]{0.48\textwidth}
    \centering
    \includegraphics[width=\textwidth]{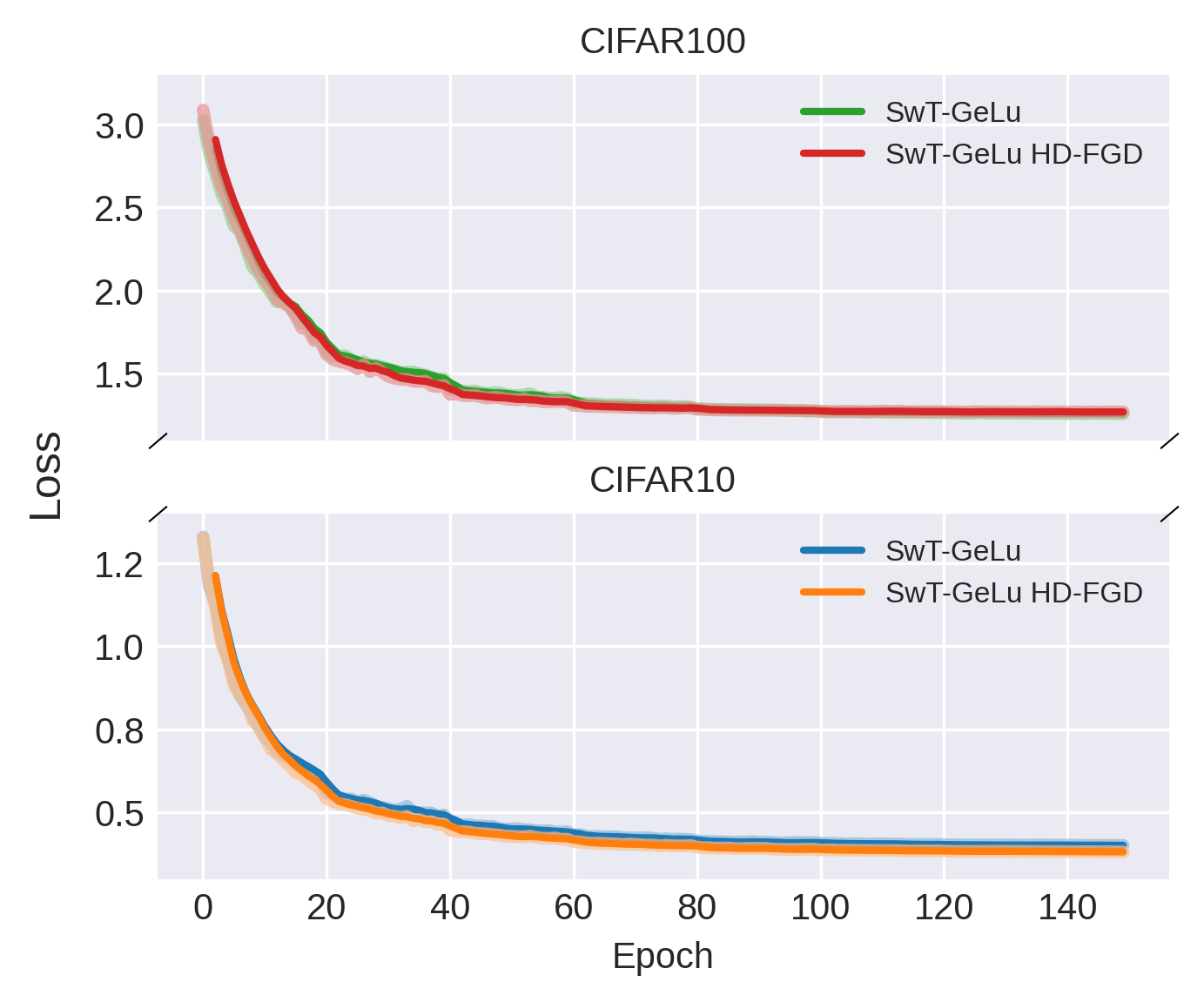}
    \subcaption{SwT + GeLu}
  \end{minipage}
  \hfill
  \begin{minipage}[b]{0.48\textwidth}
    \centering
    \includegraphics[width=\textwidth]{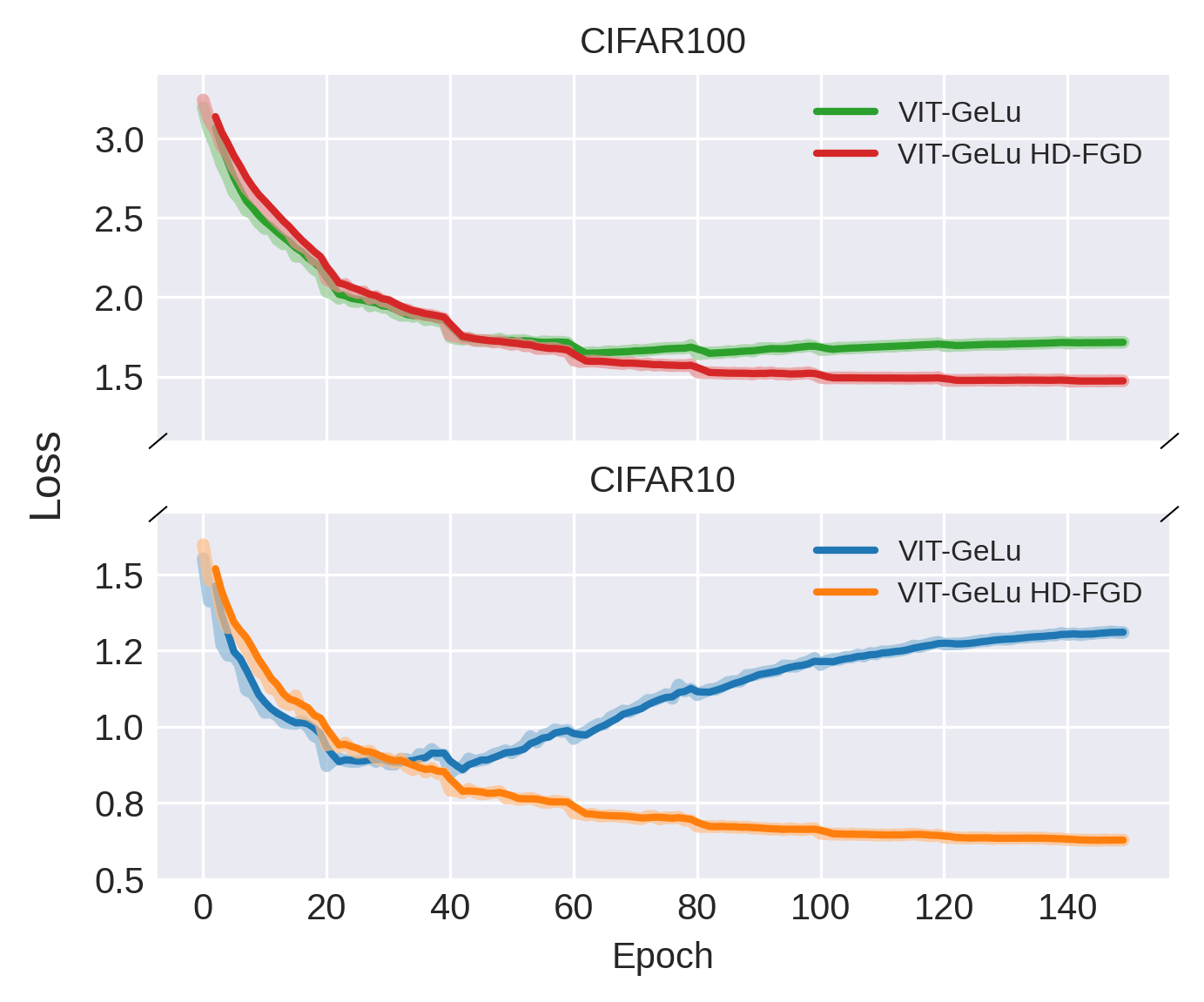}
    \subcaption{VIT + GeLu}
  \end{minipage}
  \caption{Compared with GeLu, HD-FGD test losses of GeLu in different scenario tasks.}
  \label{fig:6}
\end{figure}

Figure \ref{fig:5} displayed the loss curves on the test set for SwT and VIT models with Tanh as the activation function and the decomposed form of Tanh obtained through HD-FGD. Both curves using HD-FGD were noticeably lower than the original activation function. Similarly, the loss convergence curves for GeLu after applying HD-FGD were lower than the baseline in all models, as shown in Figure \ref{fig:6}.

\begin{table}
\centering
\caption{ACC, Top-5 Error, and Epoch Time based on HD-FGD. AF represents activation function, HD-FGD represents high-dimensional function graph decomposition.}
\begin{tabular}{ccccccc} 
\toprule
\textbf{Dataset}                   & \textbf{Model}                    & \textbf{AF}           & \textbf{HD-FGD} & \textbf{ACC}    & \textbf{Top-5 Error} & \textbf{Epoch Time}  \\ 
\midrule
\multirow{8}{*}{\raisebox{-1.5ex}[0pt][0pt]{\rotatebox[origin=c]{90}{\textbf{CIFAR10}}}}  & \multirow{4}{*}{VIT}              & \multirow{2}{*}{Tanh} & $\checkmark$ & \textbf{75.23$\pm$0.27} & \textbf{1.49$\pm$0.07}       & \textbf{89.38$\pm$0.93}      \\
                                   &                                   &                       & $\times$     & 64.64$\pm$0.28          & 2.83$\pm$0.24               & 104.55$\pm$0.14             \\ 
\cmidrule(lr){3-3}\cmidrule(lr){4-4}\cmidrule(lr){5-5}\cmidrule(lr){6-6}\cmidrule(lr){7-7}
                                   &                                   & \multirow{2}{*}{GeLu} & $\checkmark$ & \textbf{72.22$\pm$0.69} & \textbf{1.76$\pm$0.08}       & \textbf{88.11$\pm$0.18}     \\
                                   &                                   &                       & $\times$     & 64.53$\pm$0.43         & 3.27$\pm$0.19                & 140.90$\pm$0.24              \\ 
\cmidrule(r){2-7}
                                   & \multirow{4}{*}{SwT} & \multirow{2}{*}{Tanh} & $\checkmark$ & \textbf{79.00$\pm$0.28} & \textbf{1.03$\pm$0.07}       & \textbf{20.07$\pm$0.03}      \\
                                   &                                   &                       & $\times$     & 60.91$\pm$0.35         & 3.86$\pm$0.11                & 22.87$\pm$2.63               \\ 
\cmidrule(lr){3-3}\cmidrule(lr){4-4}\cmidrule(lr){5-5}\cmidrule(lr){6-6}\cmidrule(lr){7-7}
                                   &                                   & \multirow{2}{*}{GeLu} & $\checkmark$ & \textbf{83.74$\pm$0.21} & \textbf{0.64$\pm$0.02}      & \textbf{20.58$\pm$0.04}     \\
                                   &                                   &                       & $\times$     & 82.77$\pm$0.25          & 0.71$\pm$0.07                & 26.60$\pm$4.09              \\ 
\midrule
\multirow{8}{*}{\raisebox{-1.5ex}[0pt][0pt]{\rotatebox[origin=c]{90}{\textbf{CIFAR100}}}} & \multirow{4}{*}{VIT}              & \multirow{2}{*}{Tanh} & $\checkmark$ & \textbf{48.62$\pm$0.16} & \textbf{23.29$\pm$0.24}      & \textbf{91.00$\pm$0.19}  \\
                                   &                                   &                       & $\times$     & 41.36$\pm$0.44          & 27.47$\pm$0.35               & 103.14$\pm$1.41             \\ 
\cmidrule(lr){3-3}\cmidrule(lr){4-4}\cmidrule(lr){5-5}\cmidrule(lr){6-6}\cmidrule(lr){7-7}
                                   &                                   & \multirow{2}{*}{GeLu} & $\checkmark$ & \textbf{50.88$\pm$0.54} & \textbf{20.43$\pm$0.41}      & \textbf{89.85$\pm$0.10}      \\
                                   &                                   &                       & $\times$     & 47.92$\pm$0.29         & 23.51$\pm$0.18              & 102.05$\pm$0.13              \\ 
\cmidrule(r){2-7}
                                   & \multirow{4}{*}{SwT} & \multirow{2}{*}{Tanh} & $\checkmark$ & \textbf{49.45$\pm$0.22} & \textbf{20.90$\pm$0.24}      & \textbf{20.41$\pm$0.01}      \\
                                   &                                   &                       & $\times$     & 29.28$\pm$0.42          & 41.54$\pm$0.29               & 21.21$\pm$0.31               \\ 
\cmidrule(lr){3-3}\cmidrule(lr){4-4}\cmidrule(lr){5-5}\cmidrule(lr){6-6}\cmidrule(lr){7-7}
                                   &                                   & \multirow{2}{*}{GeLu} & $\checkmark$ & \textbf{57.31$\pm$0.40} & \textbf{15.65$\pm$0.18}      & \textbf{20.92$\pm$0.03}      \\
                                   &                                   &                       & $\times$    & 56.72$\pm$0.25          & 15.70$\pm$0.21               & 24.39$\pm$0.09               \\
\bottomrule
\end{tabular}
\label{table-2}
\end{table}

\subsubsection{Analysis of SA Experiment Results} \label{A-3-3}
In addition to using HD-FGD, further adversarial decomposition was applied by obtaining adversarial functions for Tanh and GeLu through network adversarial method. These adversarial functions, along with the decomposed activation functions obtained through HD-FGD, were alternately applied to the network layers of the models. 

The results in Table \ref{table-3} indicate that using HD-FGD or SA to replace MLP and combine it with the original activation function will bring greater gains to the model. In addition to the improvement in ACC and Top5 Error, the training time for each epoch of the model is also significantly reduced. Generally, HD-FGD will bring various performance improvements to the model, while the network adversarial approach brings greater gains in ACC and Top-5 Error. Although the training time may increase slightly, it still significantly shortens the model training time compared to the experimental benchmark. 

On the CIFAR10 dataset, adversarial decomposition led to ACC improvements ranging from 3.09\% to 40.72\% and Top-5 Error reductions ranging from 22.54\% to 86.01\% , and average runtime reductions per epoch ranging from 3.63\% to 29.39\% for the various models. On the CIFAR100 dataset, adversarial decomposition resulted in ACC improvements ranging from 5.45\% to 73.91\% and Top-5 Error reductions ranging from 4.76\% to 65.62\% , and average runtime reductions per epoch ranging from 1.87\% to 15.17\% for the different models.

\begin{figure}[h]
  \centering
  \begin{minipage}[b]{0.48\textwidth}
    \centering
    \includegraphics[width=\textwidth]{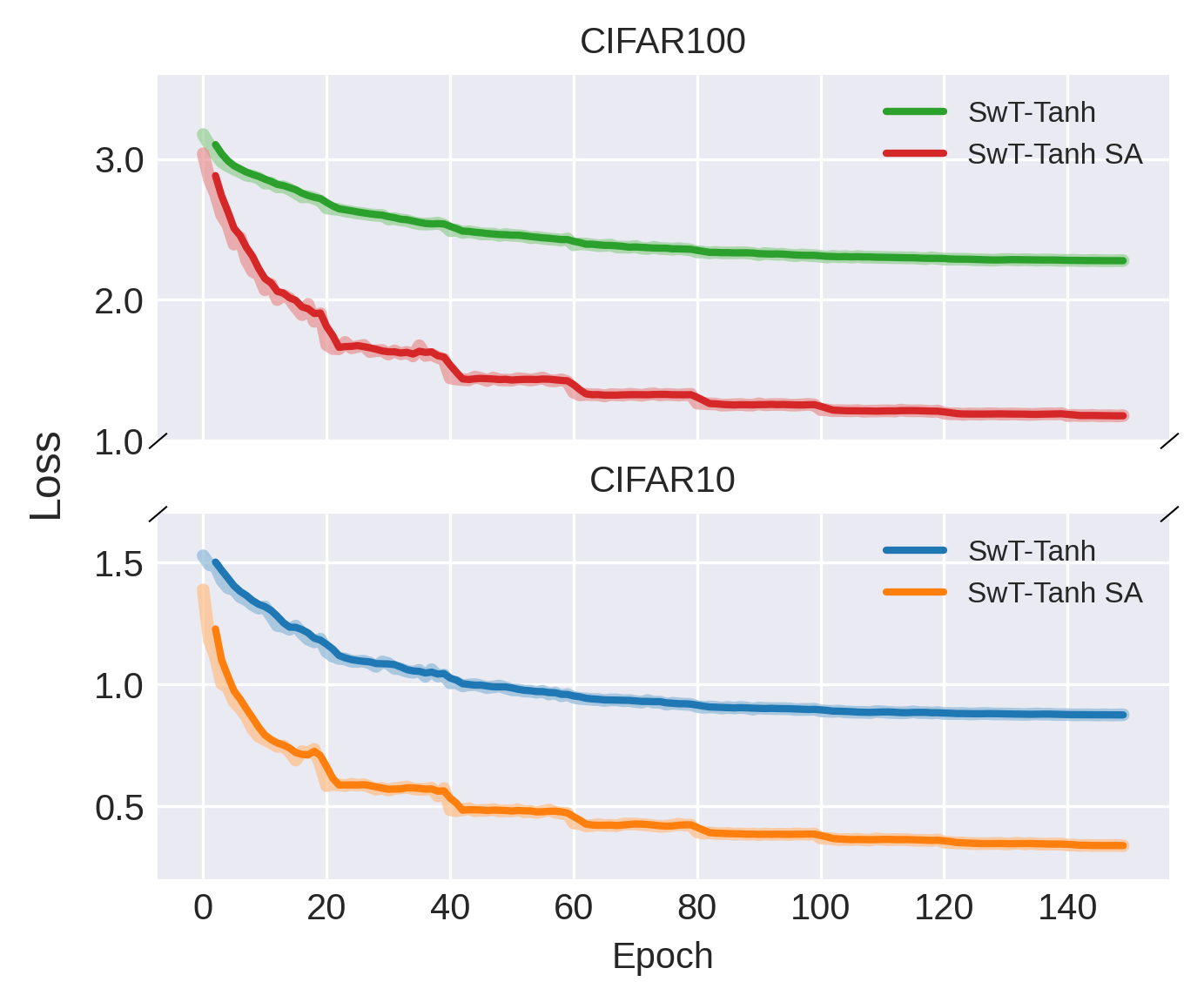}
    \subcaption{SwT + Tanh}
  \end{minipage}
  \hfill
  \begin{minipage}[b]{0.48\textwidth}
    \centering
    \includegraphics[width=\textwidth]{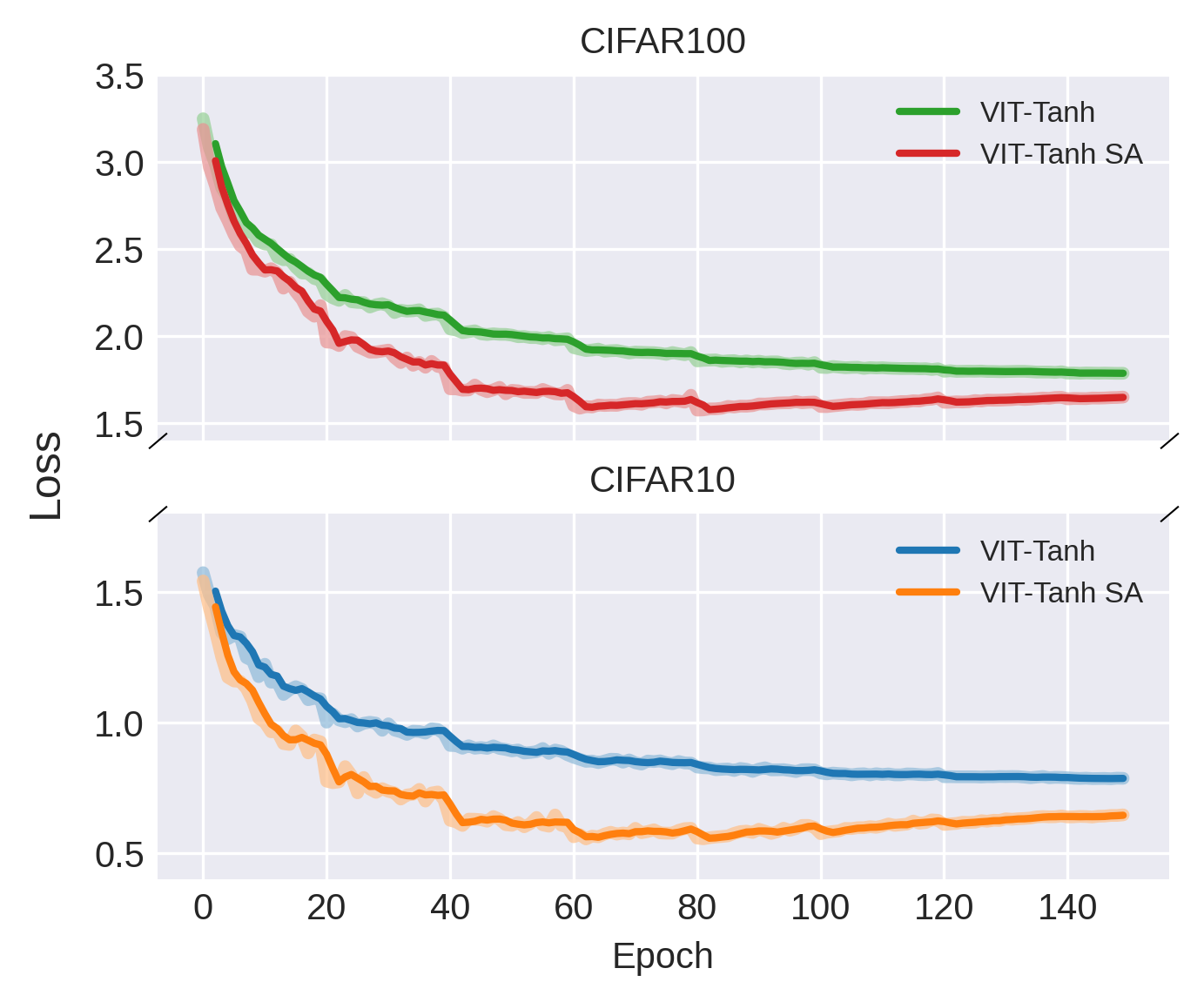}
    \subcaption{VIT + Tanh}
  \end{minipage}
  \caption{Compared with Tanh, SA test losses of Tanh in different scenario tasks.}
  \label{fig:7}
\end{figure}

\begin{figure}[h]
  \centering
  \begin{minipage}[b]{0.48\textwidth}
    \centering
    \includegraphics[width=\textwidth]{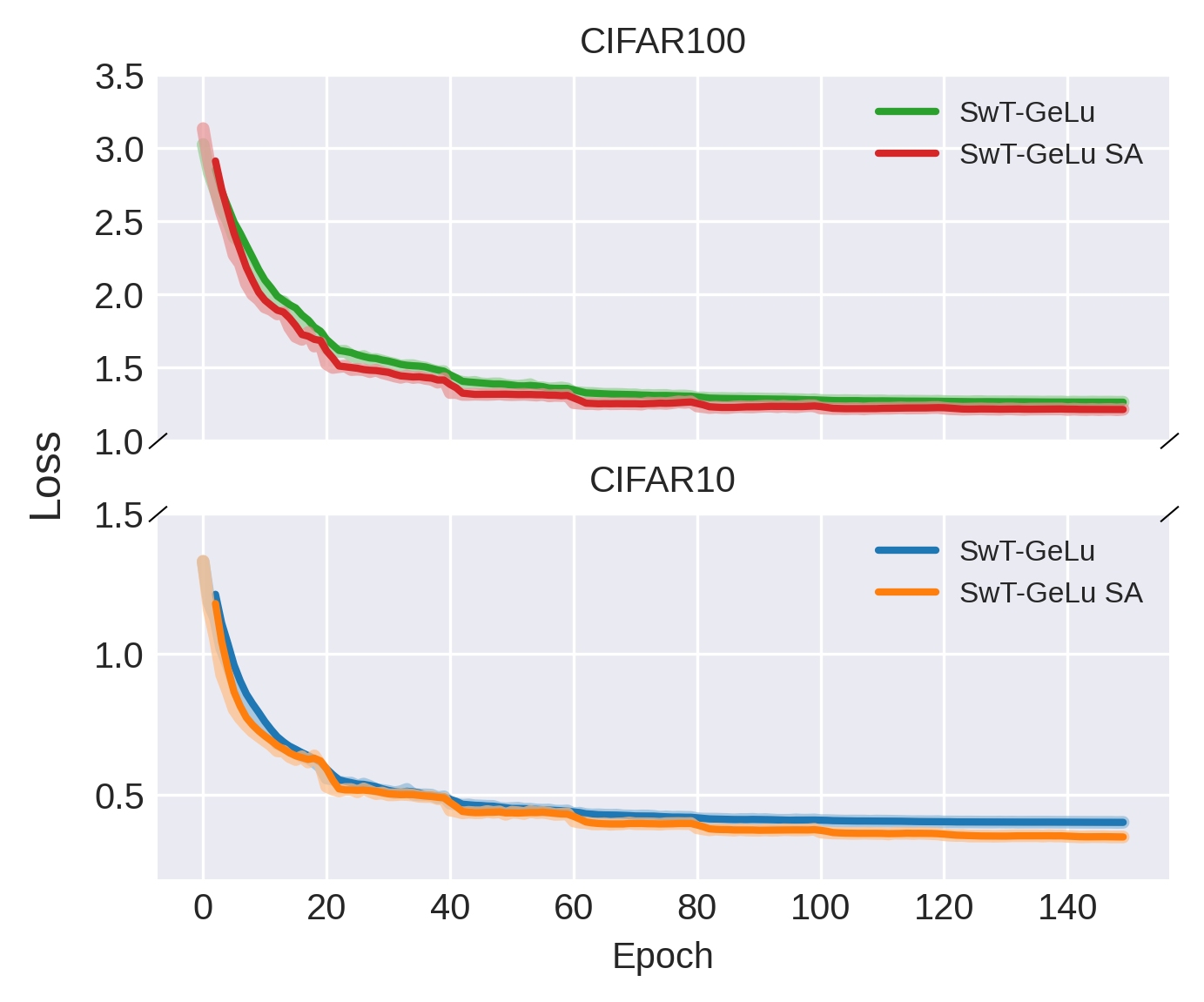}
    \subcaption{SwT + GeLu}
  \end{minipage}
  \hfill
  \begin{minipage}[b]{0.48\textwidth}
    \centering
    \includegraphics[width=\textwidth]{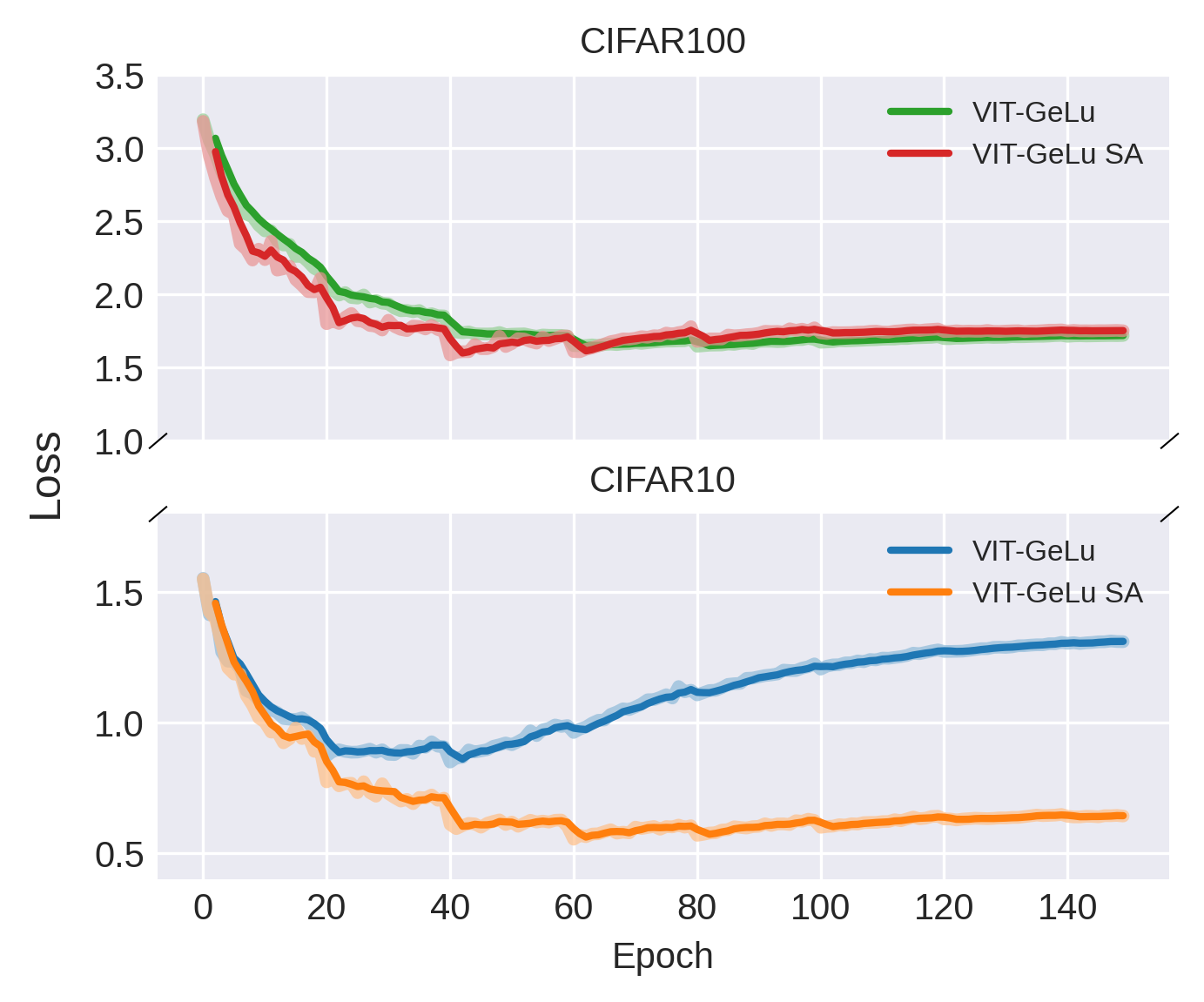}
    \subcaption{VIT + GeLu}
  \end{minipage}
  \caption{Compared with GeLu, SA test losses of GeLu in different scenario tasks.}
  \label{fig:8}
\end{figure}

Figure \ref{fig:7} and Figure \ref{fig:8} depicted the loss curves on the test set for the SwT and VIT models when using the adversarial decomposition based on Tanh and GeLu. In all experiments, the loss curve of the model using SA was significantly more stable and lower, indicating better predictive performance and faster convergence speed. It is important to note that Tanh and GeLu adopted the same decomposition form in this experiment. Exploring other decompositions based on different scenarios could potentially lead to even better performance when the task complexity increases.

\begin{table}[h]
\centering
\caption{ACC, Top-5 Error and Epoch Time based on SA. AF represents activation function, HD-FGD represents high-dimensional function graph decomposition and SA represents split adversarial.}
\begin{tabular}{cccccccc}
\toprule
\textbf{Dataset} & \textbf{Model} & \textbf{AF} & \textbf{HD-FGD} & \textbf{SA} & \textbf{ACC} & \textbf{Top-5 Error} & \textbf{Epoch Time} \\
\midrule
\multirow{12}{*}{\rotatebox[origin=c]{90}{\textbf{CIFAR10}}} & \multirow{6}{*}{VIT} & \multirow{3}{*}{Tanh} & $\checkmark$ & $\checkmark$ & \textbf{78.35}$\pm$\textbf{0.43} & \textbf{1.13}$\pm$\textbf{0.07} & 94.17$\pm$0.08 \\
 & & & $\checkmark$ & $\times$ & 75.23$\pm$0.27 & 1.49$\pm$0.07 & \textbf{89.38}$\pm$\textbf{0.93} \\
 & & & $\times$ & $\times$ & 64.64$\pm$0.28 & 2.83$\pm$0.24 & 104.54$\pm$0.14 \\
\cmidrule(lr){3-3}\cmidrule(lr){4-4}\cmidrule(lr){5-5}\cmidrule(lr){6-6}\cmidrule(lr){7-7}\cmidrule(lr){8-8}
 & & \multirow{3}{*}{GeLu} & $\checkmark$ & $\checkmark$ & \textbf{78.77}$\pm$\textbf{0.33} & \textbf{1.11}$\pm$\textbf{0.07} & 99.49$\pm$0.18 \\
 & & & $\checkmark$ & $\times$ & 72.22$\pm$0.69 & 1.76$\pm$0.08 & \textbf{88.11}$\pm$\textbf{0.18} \\
 & & & $\times$ & $\times$ & 64.53$\pm$0.43 & 3.27$\pm$0.19 & 140.90$\pm$0.24 \\
\cmidrule(lr){2-8}
 & \multirow{6}{*}{SwT} & \multirow{3}{*}{Tanh} & $\checkmark$ & $\checkmark$ & \textbf{85.71}$\pm$\textbf{0.26} & \textbf{0.54}$\pm$\textbf{0.05} & 22.04$\pm$0.03 \\
 & & & $\checkmark$ & $\times$ & 79.00$\pm$0.28 & 1.03$\pm$0.07 & \textbf{20.07}$\pm$\textbf{0.03} \\
 & & & $\times$ & $\times$ & 60.91$\pm$0.35 & 3.86$\pm$0.11 & 22.87$\pm$2.63 \\
\cmidrule(lr){3-3}\cmidrule(lr){4-4}\cmidrule(lr){5-5}\cmidrule(lr){6-6}\cmidrule(lr){7-7}\cmidrule(lr){8-8}
 & & \multirow{3}{*}{GeLu} & $\checkmark$ & $\checkmark$ & \textbf{85.33}$\pm$\textbf{0.21} & \textbf{0.55}$\pm$\textbf{0.04} & \textbf{20.35}$\pm$\textbf{0.05} \\
 & & & $\checkmark$ & $\times$ & 83.74$\pm$0.21 & 0.64$\pm$0.02 & 20.58$\pm$0.04 \\
 & & & $\times$ & $\times$ & 82.77$\pm$0.25 & 0.71$\pm$0.07 & 26.60$\pm$4.09 \\
\midrule
\multirow{12}{*}{\rotatebox[origin=c]{90}{\textbf{CIFAR100}}} & \multirow{6}{*}{VIT} & \multirow{3}{*}{Tanh} & $\checkmark$ & $\checkmark$ & \textbf{49.79}$\pm$\textbf{1.01} & \textbf{22.22}$\pm$\textbf{0.76} & 96.22$\pm$0.04 \\
 & & & $\checkmark$ & $\times$ & 48.62$\pm$0.16 & 23.29$\pm$0.24 & \textbf{91.00}$\pm$\textbf{0.19} \\
 & & & $\times$ & $\times$ & 41.36$\pm$0.44 & 27.47$\pm$0.35 & 103.14$\pm$1.41 \\
\cmidrule(lr){3-3}\cmidrule(lr){4-4}\cmidrule(lr){5-5}\cmidrule(lr){6-6}\cmidrule(lr){7-7}\cmidrule(lr){8-8}
 & & \multirow{3}{*}{GeLu} & $\checkmark$ & $\checkmark$ & \textbf{50.92}$\pm$\textbf{0.41} & 22.39$\pm$0.33 & 100.14$\pm$1.02 \\
 & & & $\checkmark$ & $\times$ & 50.88$\pm$0.54 & \textbf{20.43}$\pm$\textbf{0.41} & \textbf{89.85}$\pm$\textbf{0.10} \\
 & & & $\times$ & $\times$ & 47.92$\pm$0.29 & 23.51$\pm$0.18 & 102.05$\pm$0.13 \\
\cmidrule(lr){2-8}
 & \multirow{6}{*}{SwT} & \multirow{3}{*}{Tanh} & $\checkmark$ & $\checkmark$ & \textbf{59.92}$\pm$\textbf{0.13} & \textbf{14.28}$\pm$\textbf{0.11} & 22.35$\pm$0.03 \\
 & & & $\checkmark$ & $\times$ & 49.45$\pm$0.22 & 20.9$\pm$0.24 & \textbf{20.41}$\pm$\textbf{0.01} \\
 & & & $\times$ & $\times$ & 29.28$\pm$0.42 & 41.54$\pm$0.29 & 21.21$\pm$0.31 \\
\cmidrule(lr){3-3}\cmidrule(lr){4-4}\cmidrule(lr){5-5}\cmidrule(lr){6-6}\cmidrule(lr){7-7}\cmidrule(lr){8-8}
 & & \multirow{3}{*}{GeLu} & $\checkmark$ & $\checkmark$ & \textbf{59.81}$\pm$\textbf{0.14} & \textbf{14.91}$\pm$\textbf{0.20} & \textbf{20.69}$\pm$\textbf{0.02} \\
 & & & $\checkmark$ & $\times$ & 57.31$\pm$0.40 & 15.65$\pm$0.18 & 20.92$\pm$0.03 \\
 & & & $\times$ & $\times$ & 56.72$\pm$0.25 & 15.70$\pm$0.21 & 24.39$\pm$0.09 \\
\bottomrule
\end{tabular}
\label{table-3}
\end{table}

\subsection{Experimental Details}
In this section, we will provide the network architecture and hyperparameters used for experiments in \ref{5}. For all experiments, the following settings will be used.
\begin{itemize}
    \item Epochs: 150
    \item Batch Size: 128
    \item Optimizer: SGD
    \item Learning Rate: 0.1
    \item Learning Rate Scheduler: $\left(0.5^{ \left(\frac{\text{epoch}}
    {20}\right)} \right) \times $ Learning Rate
    \item MLP Ratio: 4
\end{itemize}
In order to maintain the same number of parameters as the benchmark model using network adversarial method, it is necessary to strictly limit the setting of the activation function splitting parameter Number of Parallel Matrices. The specific calculation method is as follows.

Assuming that the number of splitting terms for the activation function is $k$ and the Number of Parallel Matrices is $n$, then there is

\begin{equation}\label{f23}
\begin{array}{l}
 \psi _{\rm{I}}  \times \psi _{\rm{H}}  + \psi _{\rm{H}}  \times \psi _{\rm{O}}  = \psi _{\rm{I}}  \times \frac{{\psi _{\rm{H}} }}{n} \times k + \frac{{\psi _{\rm{H}} }}{n} \times \psi _{\rm{O}}  
  \Rightarrow n = \frac{{\psi _{\rm{I}}  \times \psi _{\rm{H}}  \times k + \psi _{\rm{H}}  \times \psi _{\rm{O}} }}{{\psi _{\rm{I}}  \times \psi _{\rm{H}}  + \psi _{\rm{H}}  \times \psi _{\rm{O}} }} \\ 
 \end{array}
\end{equation}
Among them, $\psi _{\rm{I}} $ represents the input feature dimension of the multi-layer perceptron, $\psi _{\rm{H}} $ represents the hidden layer feature dimension of the multi-layer perceptron, and $\psi _{\rm{O}} $ represents the output feature dimension of the multi-layer perceptron. In the VIT and SwT experiments involved in this research
\begin{equation}\label{f24}
\left\{ \begin{array}{l}
 \psi _{\rm{I}}  = {\rm{Embedding ~ Dimension}} \\ 
 \psi _{\rm{H}}  = {\rm{Embedding ~ Dimension}} \times {\rm{MLP ~ Ratio}} \\ 
 \psi _{\rm{O}}  = {\rm{Embedding ~ Dimension}} \\ 
 \end{array} \right.
\end{equation}
Among them, Embedding Dimension represents the output dimension of the embedding layer, and MLP Ratio represents the ratio of the width of the hidden layer to its input width in a multi-layer perceptron. By substituting formula (\ref{f24}) into formula (\ref{f23}), the solution is $n = \frac{{k + 1}}{2}$. Therefore, when the original activation function is split into 3 terms, $n=2$, and when the activation function is split into 4 terms,$ n=2.5$. In the experiments of this paper, $n=4$, which results in fewer parameters compared to the baseline model.

Whether to add batch normalization in the network layer is based on the decision criterion of having the best effect on the normal activation function model. For other details of different models under different scene tasks, the following settings are provided.

For \textbf{CIFAR10} dataset, common settings as follows.

Hyperparameters for \textbf{ResNet with Sigmoid} are as follows.
\begin{itemize}
    \item Activation Function: $\rm{Sigmoid(x)}$
\end{itemize}

Hyperparameters for \textbf{ResNet with Sigmoid GA} are as follows.
\begin{itemize}
    \item Activation Function: ${\rm{Sigmoid}}_\vartheta  (x)$ + $\xi _{{\rm{Sigmoid}}_\vartheta  } (x)$
\end{itemize}

For all SwT experiments, the following settings were used
\begin{itemize}
    \item Patch Size: 4
    \item Window Size: 7
    \item Embedding Dimension:96
    \item Depths: (2, 2, 6, 2)
    \item Num Heads: (3, 6, 12, 24)
\end{itemize}

Hyperparameters for \textbf{SwT with Sigmoid} are as follows.
\begin{itemize}
    \item Activation Function: $\rm{Sigmoid(x)}$
\end{itemize}

Hyperparameters for \textbf{SwT with Sigmoid GA} are as follows.
\begin{itemize}
    \item Activation Function: ${\rm{Sigmoid}}_\vartheta  (x)$ + $\xi _{{\rm{Sigmoid}}_\vartheta  } (x)$
\end{itemize}

Hyperparameters for \textbf{SwT with GeLu} are as follows.
\begin{itemize}
    \item Activation Function: $\rm{GeLu(x)}$
    \item Batch Normalization: True
\end{itemize}

Hyperparameters for \textbf{SwT with Tanh} are as follows.
\begin{itemize}
    \item Activation Function: $\rm{Tanh(x)}$
    \item Batch Normalization: True
\end{itemize}

Hyperparameters for \textbf{SwT with Tanh after HD-FGD} are as follows.
\begin{itemize}
    \item Activation Function: $\rm{Tanh(\chi _1 ,\chi _2 ,\chi _3, \chi _4 )}$
    \item SA: False
    \item Number of Parallel Matrices: 4
    \item Batch Normalization: True
\end{itemize}

Hyperparameters for \textbf{SwT with Tanh SA} are as follows.
\begin{itemize}
    \item Activation Function: $\rm{Tanh(x)}$ + $\xi _{{\rm{Tanh}}} (\chi _1 ,\chi _2 ,\chi _3 , \chi _4 )$
    \item SA: True
    \item Number of Parallel Matrices: 4
    \item L2 Penalty Coefficient: 0.05
    \item Batch Normalization: True
\end{itemize}

Hyperparameters for \textbf{SwT with GeLu after HD-FGD} are as follows.
\begin{itemize}
    \item Activation Function: ${\rm{GeLu}}(\chi _1 ,\chi _2 ,\chi _3, \chi _4 )$
    \item SA: False
    \item Number of Parallel Matrices: 4
    \item L2 Penalty Coefficient: 0.005
    \item Batch Normalization: True
\end{itemize}

Hyperparameters for \textbf{SwT with GeLu SA} are as follows.
\begin{itemize}
    \item Activation Function: ${\rm{GeLu}}(\chi _1 ,\chi _2 ,\chi _3, \chi _4 )$
    \item SA: True
    \item Number of Parallel Matrices: 4
    \item L2 Penalty Coefficient: 0.05
    \item Batch Normalization: True
\end{itemize}

For all VIT experiments, the following settings were used
\begin{itemize}
    \item Patch Size: 4
    \item Embedding Dimension:768
    \item Depth: 12
    \item Num Heads: 12
\end{itemize}

Hyperparameters for \textbf{VIT with Sigmoid} are as follows.
\begin{itemize}
    \item Activation Function: $\rm{Sigmoid(x)}$
\end{itemize}

Hyperparameters for \textbf{VIT with Sigmoid GA} are as follows.
\begin{itemize}
    \item Activation Function: ${\rm{Sigmoid}}_\vartheta  (x)$ + $\xi _{{\rm{Sigmoid}}_\vartheta  } (x)$
\end{itemize}

Hyperparameters for \textbf{VIT with GeLu} are as follows.
\begin{itemize}
    \item Activation Function: $\rm{GeLu(x)}$
\end{itemize}

Hyperparameters for \textbf{VIT with Tanh} are as follows.
\begin{itemize}
    \item Activation Function: $\rm{Tanh(x)}$
\end{itemize}

Hyperparameters for \textbf{VIT with Tanh after HD-FGD} are as follows.
\begin{itemize}
    \item Activation Function: $\rm{Tanh(\chi _1 ,\chi _2 ,\chi _3 ,\chi _4 )}$
    \item SA: False
    \item Number of Parallel Matrices: 4
\end{itemize}

Hyperparameters for \textbf{VIT with Tanh SA} are as follows.
\begin{itemize}
    \item Activation Function: $\rm{Tanh(x)}$ + $\xi _{{\rm{Tanh}}} (\chi _1 ,\chi _2 ,\chi _3,\chi _4 )$
    \item SA: True
    \item Number of Parallel Matrices: 4
    \item L2 Penalty Coefficient: 0.05
\end{itemize}

Hyperparameters for \textbf{VIT with GeLu after HD-FGD} are as follows.
\begin{itemize}
    \item Activation Function: ${\rm{GeLu}}(\chi _1 ,\chi _2 ,\chi _3 ,\chi _4 )$
    \item SA: False
    \item Number of Parallel Matrices: 4
\end{itemize}

Hyperparameters for \textbf{VIT with GeLu SA} are as follows.
\begin{itemize}
    \item Activation Function: ${\rm{GeLu}}(\chi _1 ,\chi _2 ,\chi _3 ,\chi _4 )$
    \item SA: True
    \item Number of Parallel Matrices: 4
    \item L2 Penalty Coefficient: 0.05
\end{itemize}

For \textbf{CIFAR100} dataset, common settings as follows.

Hyperparameters for \textbf{ResNet with Sigmoid} are as follows.
\begin{itemize}
    \item Activation Function: $\rm{Sigmoid(x)}$
\end{itemize}

Hyperparameters for \textbf{ResNet with Sigmoid GA} are as follows.
\begin{itemize}
    \item Activation Function: ${\rm{Sigmoid}}_\vartheta  (x)$ + $\xi _{{\rm{Sigmoid}}_\vartheta  } (x)$
\end{itemize}

For all SwT experiments, the following settings were used
\begin{itemize}
    \item Patch Size: 4
    \item Window Size: 7
    \item Embedding Dimension:96
    \item Depths: (2, 2, 6, 2)
    \item Num Heads: (3, 6, 12, 24)
\end{itemize}

Hyperparameters for \textbf{SwT with Sigmoid} are as follows.
\begin{itemize}
    \item Activation Function: $\rm{Sigmoid(x)}$
\end{itemize}

Hyperparameters for \textbf{SwT with Sigmoid GA} are as follows.
\begin{itemize}
    \item Activation Function: ${\rm{Sigmoid}}_\vartheta  (x)$ + $\xi _{{\rm{Sigmoid}}_\vartheta  } (x)$
\end{itemize}

Hyperparameters for \textbf{SwT with GeLu} are as follows.
\begin{itemize}
    \item Activation Function: $\rm{GeLu(x)}$
    \item Batch Normalization: True
\end{itemize}

Hyperparameters for \textbf{SwT with Tanh} are as follows.
\begin{itemize}
    \item Activation Function: $\rm{Tanh(x)}$
    \item Batch Normalization: True
\end{itemize}

Hyperparameters for \textbf{SwT with Tanh after HD-FGD} are as follows.
\begin{itemize}
    \item Activation Function: $\rm{Tanh(\chi _1 ,\chi _2 ,\chi _3,\chi _4 )}$
    \item SA: False
    \item Number of Parallel Matrices: 4
    \item Batch Normalization: True
\end{itemize}

Hyperparameters for \textbf{SwT with Tanh SA} are as follows.
\begin{itemize}
    \item Activation Function: $\rm{Tanh(x)}$ + $\xi _{{\rm{Tanh}}} (\chi _1 ,\chi _2 ,\chi _3 ,\chi _4 )$
    \item SA: True
    \item Number of Parallel Matrices: 4
    \item L2 Penalty Coefficient: 0.05
    \item Batch Normalization: True
\end{itemize}

Hyperparameters for \textbf{SwT with GeLu after HD-FGD} are as follows.
\begin{itemize}
    \item Activation Function: ${\rm{GeLu}}(\chi _1 ,\chi _2 ,\chi _3 )$
    \item SA: False
    \item Number of Parallel Matrices: 3
    \item L2 Penalty Coefficient: 0.005
    \item Batch Normalization: True
\end{itemize}

Hyperparameters for \textbf{SwT with GeLu SA} are as follows.
\begin{itemize}
    \item Activation Function: ${\rm{GeLu}}(\chi _1 ,\chi _2 ,\chi _3,\chi _4 )$
    \item SA: True
    \item Number of Parallel Matrices: 4
    \item L2 Penalty Coefficient: 0.05
    \item Batch Normalization: True
\end{itemize}

For all VIT experiments, the following settings were used
\begin{itemize}
    \item Patch Size: 4
    \item Embedding Dimension:768
    \item Depth: 12
    \item Num Heads: 12
\end{itemize}

Hyperparameters for \textbf{VIT with Sigmoid} are as follows.
\begin{itemize}
    \item Activation Function: $\rm{Sigmoid(x)}$
\end{itemize}

Hyperparameters for \textbf{VIT with Sigmoid GA} are as follows.
\begin{itemize}
    \item Activation Function: ${\rm{Sigmoid}}_\vartheta  (x)$ + $\xi _{{\rm{Sigmoid}}_\vartheta  } (x)$
\end{itemize}

Hyperparameters for \textbf{VIT with GeLu} are as follows.
\begin{itemize}
    \item Activation Function: $\rm{GeLu(x)}$
\end{itemize}

Hyperparameters for \textbf{VIT with Tanh} are as follows.
\begin{itemize}
    \item Activation Function: $\rm{Tanh(x)}$
\end{itemize}

Hyperparameters for \textbf{VIT with Tanh after HD-FGD} are as follows.
\begin{itemize}
    \item Activation Function: $\rm{Tanh(\chi _1 ,\chi _2 ,\chi _3, \chi _4 )}$
    \item SA: False
    \item Number of Parallel Matrices: 4
\end{itemize}

Hyperparameters for \textbf{VIT with Tanh SA} are as follows.
\begin{itemize}
    \item Activation Function: $\rm{Tanh(x)}$ + $\xi _{{\rm{Tanh}}} (\chi _1 ,\chi _2 ,\chi _3 ,\chi _4 )$
    \item SA: True
    \item Number of Parallel Matrices: 4
    \item L2 Penalty Coefficient: 0.15
\end{itemize}

Hyperparameters for \textbf{VIT with GeLu after HD-FGD} are as follows.
\begin{itemize}
    \item Activation Function: ${\rm{GeLu}}(\chi _1 ,\chi _2 ,\chi _3,\chi _4 )$
    \item SA: False
    \item Number of Parallel Matrices: 4
\end{itemize}

Hyperparameters for \textbf{VIT with GeLu SA} are as follows.
\begin{itemize}
    \item Activation Function: ${\rm{GeLu}}(\chi _1 ,\chi _2 ,\chi _3 ,\chi _4)$
    \item SA: True
    \item Number of Parallel Matrices: 4
    \item L2 Penalty Coefficient: 0.08
\end{itemize}

\subsection{Adversarial Method in the Field of NLP Have Been Proven to be Effective} \label{A-4}

Translation of network adversarial method to test their effectiveness in the NLP field involves experiments using an LSTM-based autoencoder network with Sigmoid activation functions to create a baseline Seq2Seq model. In these experiments, the number of epochs is set to 50, the learning rate to 0.001, and the batch size to 128. Both the encoder and decoder have an embedding size of 300, and the output factor for both is 0.5. Training is conducted using the Adam optimizer for cross-entropy loss. We replicated some experiments from Table 11 in reference \cite{dubey2022activation}, with all experimental results averaging over five random seeds. Before the dropout layer, the activation function is applied to the feature embeddings. Experimental results in \ref{table-4} show that the Sigmoid activation function resulted in the poorest translation performance, while GeLu yielded the best translation results. By applying adversarial method to the network, using its adversarial function as the activation function in the LSTM's Encoder and Decoder, the final model's translation performance was better than that using GeLu as the activation function.

\begin{table}[h]
\centering
\caption{Experimental results for German to English language translation.}
\begin{tabular}{ccc} 
\toprule
\multicolumn{2}{c}{\textbf{~Activation}}                  & \multirow{2}{*}{\textbf{Bleu Score}}  \\ 
\cmidrule(lr){1-2}
\multicolumn{1}{l}{Encoder} & \multicolumn{1}{l}{Decoder} &                                       \\ 
\hline
Sigmoid                     & Sigmoid                     & 14.57$\pm$0.83                           \\
$\xi _{{\rm{Sigmoid}}} \left( x \right)$                     & $\xi _{{\rm{Sigmoid}}} \left( x \right)$                     & \textbf{21.88}$\pm$\textbf{0.35}                           \\
GeLu                        & GeLu                        & 21.30$\pm$0.29                           \\
ReLu                        & ReLU                        & 20.81$\pm$0.34                           \\
Tanh                        & Tanh                        & 21.10$\pm$0.40                           \\
\bottomrule
\end{tabular}
\label{table-4}
\end{table}

This demonstrates the potential of network adversarial methods in the NLP field, and we look forward to exploring its effectiveness in more areas and believe in further advancements in the future.

\end{document}